\documentclass{article}

\usepackage{amsmath}
\usepackage{template/PRIMEarxiv}

\usepackage{footmisc}

\usepackage{graphicx} 
\usepackage{tikz}
\usepackage{forest}
\usetikzlibrary{trees,positioning,shapes,shadows,arrows.meta}

\usepackage[utf8]{inputenc} 
\usepackage[T1]{fontenc}    
\usepackage{hyperref}       
\usepackage{url}            
\usepackage{booktabs}       
\usepackage{amsfonts}       
\usepackage{nicefrac}       
\usepackage{microtype}      
\usepackage{lipsum}
\usepackage{fancyhdr}       
\usepackage{graphicx}       
\graphicspath{{media/}}     

\title{Surveying the MLLM Landscape: A Meta-Review of Current Surveys}

\author{
    Ming Li\thanks{Equal contribution, Georgia Institute of Technology, mli694@gatech.edu} \and 
    Keyu Chen\thanks{Equal contribution, Georgia Institute of Technology, kchen637@gatech.edu} \and 
    Ziqian Bi\thanks{Equal contribution, Indiana University, bizi@iu.edu} \and 
    Ming Liu\thanks{ Purdue University, liu3183@purdue.edu} \and 
    Xinyuan Song\thanks{Equal contribution, Emory University, xsong30@emory.edu} \and
    Zekun Jiang\thanks{Equal contribution, Sichuan University, zekun\_jiang@163.com} \and
    Tianyang Wang\thanks{Equal contribution, Xi’an Jiaotong-Liverpool University, Tianyang.Wang21@student.xjtlu.edu.cn} \and   
    Benji Peng\thanks{AppCubic, benji@appcubic.com} \and 
    Qian Niu\thanks{Kyoto University, niuqian1995@gmail.com} \and 
    Junyu Liu\thanks{Kyoto University, juniorliu95@gmail.com} \and 
    Jinlang Wang\thanks{University of Wisconsin-Madison, jinlang.wang@wisc.edu} \and 
    Sen Zhang\thanks{Rutgers University, sen.z@rutgers.edu} \and 
    Xuanhe Pan\thanks{University of Wisconsin-Madison, xpan73@wisc.edu} \and 
    Jiawei Xu\thanks{Purdue University, xu1644@purdue.edu} \and 
    Pohsun Feng\thanks{Corresponding author, National Taiwan Normal University, 41075018h@ntnu.edu.tw} 
}

\begin{document}

\maketitle

\begin{abstract}
The rise of Multimodal Large Language Models (MLLMs) has become a transformative force in the field of artificial intelligence, enabling machines to process and generate content across multiple modalities, such as text, images, audio, and video. These models represent a significant advancement over traditional unimodal systems, opening new frontiers in diverse applications ranging from autonomous agents to medical diagnostics. By integrating multiple modalities, MLLMs achieve a more holistic understanding of information, closely mimicking human perception. As the capabilities of MLLMs expand, the need for comprehensive and accurate performance evaluation has become increasingly critical. This survey aims to provide a systematic review of benchmark tests and evaluation methods for MLLMs, covering key topics such as foundational concepts, applications, evaluation methodologies, ethical concerns, security, efficiency, and domain-specific applications. Through the classification and analysis of existing literature, we summarize the main contributions and methodologies of various surveys, conduct a detailed comparative analysis, and examine their impact within the academic community. Additionally, we identify emerging trends and underexplored areas in MLLM research, proposing potential directions for future studies. This survey is intended to offer researchers and practitioners a comprehensive understanding of the current state of MLLM evaluation, thereby facilitating further progress in this rapidly evolving field.
\end{abstract}


\newpage
\tableofcontents 

\section{Introduction}

\subsection{Purpose of the Survey}

Multimodal Large Language Models (MLLMs) represent a significant advancement in artificial intelligence, allowing systems to process and generate content across diverse modalities, such as text, images, audio, and video. By integrating multiple data types, MLLMs move beyond the limitations of unimodal models, enabling more comprehensive and sophisticated applications in areas ranging from autonomous systems to medical diagnostics.

Given the rapid development of MLLMs, the field has produced a wealth of surveys, each exploring specific aspects of these models. However, the sheer volume and diversity of this literature can make it difficult for researchers and practitioners to grasp the current state of the field. To address this, we present a ``survey of surveys'' that synthesizes key insights across existing reviews and organizes them into 11 core areas: General, Evaluation, Security, Bias, Agents, Applications, Retrieval-Augmented Generation (RAG), Graphs, Data, Continual Learning, and Efficient Learning.

This paper aims to: \begin{itemize} \item Synthesize and categorize findings from various surveys, providing a structured overview of the most critical advancements in MLLM research. \item Identify major themes, trends, and challenges in MLLM evaluation, highlighting benchmarks, datasets, and performance metrics. \item Examine current methodologies for MLLM application and assessment, identifying gaps and suggesting improvements. \item Highlight future research directions and underexplored areas within the MLLM landscape. \end{itemize}

By offering a comprehensive synthesis of existing literature, this "survey of surveys" serves as a valuable resource for navigating the evolving MLLM field, fostering deeper understanding and guiding future research.

\section{Classical Natural Language Processing Methods}

The development of Natural Language Processing (NLP) has undergone multiple stages including statistical and rule-based methods, neural networks, word embeddings, deep word embeddings, attention mechanisms, and context-aware models. Each stage has significant technological advancements and the introduction of important models. Early research mainly focused on statistical and rule-based methods, which were later followed by the introduction of neural networks and deep learning concepts, greatly enhancing the performance and application range of NLP models.

Early language model research was primarily based on the Markov process. Markov chains were used to describe random processes and sequence data, introducing basic methods for handling sequential data\cite{dynkin1965markov, chung1967markov}. Subsequently, Hidden Markov Models (HMMs) were proposed to perform continuous speech recognition via maximum likelihood estimation, dealing with hidden states in sequential data\cite{bahl1983maximum}.
In the 1990s, n-gram-based statistical language models were widely used. N-gram models predicted the next word based on the context, establishing a simple yet effective foundation for language models\cite{jurafsky2000speech}. With increased computational power, neural networks began to be applied in NLP.
In 1997, Long Short-Term Memory (LSTM) networks were proposed to solve the problem of long-term dependencies using memory cells, making RNNs more effective in handling long sequential data\cite{grossberg2013recurrent, hochreiter1997long}.
In 1998, Yann LeCun and others further developed RNNs for document recognition, capable of handling sequential data\cite{lecun1998gradient}.
Word embedding techniques achieved significant breakthroughs in the early 2010s. In 2003, the Word2Vec model was introduced, capturing semantic relationships between words through efficient word vector training methods\cite{mikolov2013efficient}. This method introduced the Skip-gram and CBOW models, greatly improving the quality of word vectors.

In 2009, ImageNet, a large visual database containing over 14 million annotated images, was launched for object recognition and image classification tasks\cite{deng2009imagenet}. The advent of ImageNet provided deep learning with abundant data resources, making it possible to train deep neural networks on large-scale datasets.
In 2012, AlexNet's success in the ImageNet\cite{deng2009imagenet} competition marked the resurgence of deep learning\cite{krizhevsky2012imagenet}. AlexNet significantly improved image classification performance through deep convolutional neural networks, bringing deep learning back to the forefront of research and inspiring researchers to explore its potential in other fields like NLP.
In 2014, the GloVe model trained word vectors using global word co-occurrence matrices, further improving semantic representation of words\cite{pennington2014glove}. Additionally, the sequence-to-sequence learning (Seq2Seq) model was proposed for machine translation, handling sequence transformation tasks through an encoder-decoder structure\cite{sutskever2014sequence}.
In 2015, Bahdanau et al. introduced the attention mechanism, which dynamically aligned source and target sequences, improving machine translation performance\cite{bahdanau2014neural}.
In 2016, the Byte Pair Encoding (BPE) algorithm was proposed to handle rare and out-of-vocabulary words through subword units, significantly enhancing model generalization ability\cite{sennrich2015neural}.
In 2018, the ELMo (Embeddings from Language Models) model was introduced, capturing contextual information through bidirectional LSTM (Long Short-Term Memory) networks. Specifically, ELMo used two independently trained LSTMs, one processing text from left to right and the other from right to left, concatenating the hidden states of these two LSTMs to form the final word representation. This method improved the quality of word representations by capturing lexical semantics and syntactic information while fully utilizing contextual information\cite{peters2018deep}.

Early language models, which relied on simple statistical methods, difficult-to-train and scale long short-term memory networks (LSTMs), or neural networks with smaller parameter sizes, often produced unsatisfactory results. These models struggled to handle the diversity and complexity of language, particularly in understanding synonyms or words in different contexts. Simple statistical methods typically considered only word frequency and order, lacking a deep understanding of semantics and context. While LSTMs and small-scale neural networks improved the model's memory capacity and ability to handle complexity to some extent, their performance gains were limited due to the complexity of the training process and computational constraints. These limitations made early language models inadequate for handling complex natural language processing tasks in practical applications.
\section{Large Language Models (LLM) and Multimodal Large Language Models (MLLM)}

\subsection{Transformer, BERT, and GPT}

Transformer, BERT, and GPT are the foundation models of large language models. These models play a crucial role in the field of natural language processing (NLP), driving the development of numerous applications and research advancements.

\subsection{Transformer}

Vaswani et al. introduced the Transformer, a deep learning model widely used for natural language processing (NLP) tasks \cite{vaswani2017attention}. As the cornerstone of large-scale language models (LLMs) today, the core of the Transformer model is the attention mechanism, which captures relationships between different positions in the input sequence without relying on traditional recurrent neural network (RNN) structures. This enables parallel computation and improves training efficiency.

The Transformer model consists of an encoder and a decoder, each composed of multiple identical layers. Each layer in the encoder contains two sub-layers: a multi-head self-attention mechanism and a feed-forward neural network. Each layer in the decoder, however, includes three sub-layers: a self-attention mechanism, an encoder-decoder attention mechanism, and a feed-forward neural network. The encoder maps the input sequence to a continuous representation, while the decoder generates the output sequence based on this representation.

\paragraph{Self-Attention Mechanism}

The self-attention mechanism operates on three matrices: the query matrix \( Q \), the key matrix \( K \), and the value matrix \( V \). The computation is performed as follows:

\[
\text{Attention}(Q, K, V) = \text{softmax}\left(\frac{QK^T}{\sqrt{d_k}}\right)V,
\]

where \( d_k \) is the dimension of the key matrix.

\paragraph{Multi-Head Attention Mechanism}

The multi-head attention mechanism computes multiple self-attention heads in parallel and concatenates the results:

\[
\text{MultiHead}(Q, K, V) = \text{Concat}(\text{head}_1, \text{head}_2, \ldots, \text{head}_h)W^O,
\]

Each head is computed as:

\[
\text{head}_i = \text{Attention}(QW_i^Q, KW_i^K, VW_i^V),
\]

where \( W_i^Q \), \( W_i^K \), \( W_i^V \), and \( W^O \) are trainable weight matrices.

\paragraph{Feed-Forward Neural Network}

The feed-forward neural network within each layer consists of two linear transformations and a ReLU activation function:

\[
\text{FFN}(x) = \text{ReLU}(xW_1 + b_1)W_2 + b_2,
\]

Here, \( W_1 \), \( W_2 \), \( b_1 \), and \( b_2 \) are trainable weights and biases. This network helps in transforming the input data into a more abstract representation before passing it to the next layer.

\subsubsection{BERT and GPT}

The encoder of the Transformer, when used independently and after undergoing pre-training and fine-tuning, becomes BERT; similarly, the decoder, after similar pre-training and fine-tuning, becomes GPT. BERT and GPT are applied to different natural language processing tasks: BERT is primarily used for understanding tasks, while GPT is mainly used for generation tasks.

BERT (Bidirectional Encoder Representations from Transformers) is a pre-training model proposed by Devlin et al.\cite{devlin2018bert}. The core idea of BERT is to generate contextual representations of words using a bidirectional Transformer encoder. It achieves performance improvements across various NLP tasks by performing unsupervised pre-training on large-scale corpora, followed by fine-tuning on specific tasks. BERT's pre-training involves two tasks: Masked Language Model (MLM) and Next Sentence Prediction (NSP).

In the MLM task, BERT randomly masks some words in the input text and predicts these masked words using the model, thereby learning the relationships between words. In the NSP task, BERT learns the relationships between sentences by determining whether two sentences are continuous.

GPT (Generative Pre-trained Transformer) is a series of generative pre-training models proposed by OpenAI\cite{radford2018improving, radford2019language, brown2020language}. Unlike BERT, GPT mainly uses a unidirectional Transformer decoder to generate text. GPT's training consists of two stages: first, pre-training on large-scale unlabelled data, and then fine-tuning on specific tasks. During the pre-training stage, the model uses an autoregressive method to predict the next word, thereby learning the sequence information of words.

GPT performs exceptionally well in generation tasks because it can generate coherent and meaningful text based on the given context. With the development of the GPT series, GPT-2 and GPT-3 introduced more parameters and larger model scales, further improving generation quality and task generalization capabilities.

\subsection{Multimodal Large Language Models (MLLM)}

Multimodal Large Language Models (MLLM) represent a significant advancement in the field of artificial intelligence by integrating multiple types of data—such as text, images, and audio—into a unified framework. Unlike traditional models that operate on a single modality, MLLMs are designed to understand and generate content across different modalities, enhancing their versatility and applicability to a wider range of tasks. The core challenge in developing MLLMs is achieving effective modality alignment, which involves mapping different types of data into a common representation space. This alignment allows the model to seamlessly interpret and interrelate information from various sources, thereby improving performance on tasks like image captioning, visual question answering, and multimodal translation. As a result, MLLMs hold the promise of enabling more natural and comprehensive human-computer interactions, paving the way for innovations in areas such as virtual assistants, content creation, and accessible technologies.

In the context of aligning text and images, the initial challenge was the necessity of vast amounts of data to ensure models could effectively understand and correlate different modalities. Early methods relied on manually labeled data, which was both costly and inefficient. Some of these early methods include DeViSE (A Deep Visual-Semantic Embedding Model) \cite{frome2013devise}, VSE (Visual Semantic Embeddings) \cite{kiros2014unifying} and CCA (Canonical Correlation Analysis) \cite{hardoon2004canonical}. These models attempted to align images and texts in a shared representation space using labeled datasets, but their performance was limited due to the restricted size of the data.

\subsubsection{Contrastive Learning-based Multimodal Alignment}

With the advent of contrastive learning, this field has significant progress. The core idea of contrastive learning is to bring similar sample pairs (such as related image and text pairs) closer together while pushing dissimilar pairs apart in the shared representation space, thus achieving multimodal alignment. Early contrastive learning methods focused primarily on the visual domain, such as MoCo (Momentum Contrast) \cite{he2020momentum, chen2020improved}, SimCLR (Simple Framework for Contrastive Learning of Visual Representations) \cite{chen2020simple, chen2020big}, Instance Discrimination \cite{wu2018unsupervised}, CPC (Contrastive Predictive Coding) \cite{oord2018representation}, CMC (Contrastive Multiview Coding) \cite{tian2020contrastive}, PIRL (Pretext-Invariant Representation Learning) \cite{misra2020self}, SwAV (Swapping Assignments between Multiple Views of the Same Image) \cite{caron2020unsupervised}, and BYOL (Bootstrap Your Own Latent) \cite{grill2020bootstrap}. These methods leveraged large unlabeled image datasets and used contrastive loss functions to significantly enhance image feature representations.

In the realm of multimodal data alignment, contrastive learning methods have made remarkable strides. CLIP (Contrastive Language-Image Pre-Training) \cite{radford2021learning} represents a major breakthrough in this area. CLIP leverages a large-scale dataset of image-text pairs and employs a contrastive loss function to achieve robust text-image alignment. Specifically, CLIP encodes images and texts into a shared representation space to align them. Given an image \( I \) and its corresponding text description \( T \), the image encoder \( f(\cdot) \) and text encoder \( g(\cdot) \) map them into a common representation space: the image feature vector \( f(I) \) and the text feature vector \( g(T) \). In this space, matched image-text pairs exhibit high similarity, while mismatched pairs show low similarity. Pretraining on hundreds of millions of image-text pairs allows CLIP to learn rich cross-modal representations, excelling in downstream tasks.

CLIP possesses strong zero-shot learning capabilities, meaning it can be directly used for feature extraction and applications without task-specific fine-tuning. For instance, in image classification, CLIP can classify images using its pretrained representations without additional labeled data. This ability makes CLIP highly flexible and efficient in practical applications, quickly adapting to new tasks and datasets. CLIP's text and image encoders can be used independently; inputting text or images into the respective encoders yields high-dimensional space features (embeddings). Specifically, the text encoder converts input text into text feature vectors \( g(T) \), and the image encoder converts input images into image feature vectors \( f(I) \), which can be utilized in various downstream tasks such as feature extraction and representation learning.

There are numerous models that employ CLIP visual or text encoders, demonstrating their broad applicability in multimodal alignment. For instance, METER (Multimodal End-to-end TransformER) \cite{dou2022empirical} integrates CLIP encoders to align image and text representations and employs optimal transport techniques to improve the performance of vision-language tasks. SimVLM (Simple Visual Language Model) \cite{wang2021simvlm} uses CLIP encoders for weakly supervised pretraining, achieving excellent performance in vision-language tasks. BLIP-2 (Bootstrapping Language-Image Pre-training) \cite{li2023blip} utilizes frozen CLIP image encoders with large language models to improve image-text pretraining. The Stable Diffusion series, including high-resolution image synthesis \cite{rombach2022high}, personalized stylization \cite{yang2023pixel}, and improved high-resolution image synthesis \cite{podell2023sdxl}, also employs CLIP encoders to optimize image generation and super-resolution tasks.

\subsubsection{Model-Based Techniques for Multimodal Alignment}

Model-based techniques for multimodal alignment aim to create unified representations of different modalities, such as images and text, by leveraging large-scale models that handle both vision and language tasks. ALIGN (Large-scale Image-Language Pretraining) \cite{DBLP:journals/corr/abs-2102-05918} employs a dual-encoder architecture that aligns images and text in a shared embedding space using contrastive learning, significantly improving zero-shot image classification and retrieval. Similarly, Florence \cite{DBLP:journals/corr/abs-2111-11432} incorporates a vision transformer model with a text encoder to align image and text modalities, demonstrating superior performance in visual understanding tasks. These models focus on pretraining using vast amounts of weakly labeled image-text data, contributing to better generalization across a range of vision-language benchmarks.

Other models utilize fine-grained alignment strategies to enhance performance in more specialized tasks. UNITER (UNiversal Image-TExt Representation) \cite{DBLP:journals/corr/abs-1909-11740} jointly learns image-text alignments using a transformer-based architecture, introducing a unique masked language modeling task to improve contextual understanding between modalities. On the other hand, ViLT (Vision-and-Language Transformer) \cite{kim2021viltvisionandlanguagetransformerconvolution} reduces the complexity of multimodal fusion by directly feeding visual patch embeddings into a transformer, bypassing the need for separate image encoders, while still maintaining competitive performance in tasks like image-text retrieval and visual question answering. These model-based techniques highlight the versatility and scalability of transformer-based approaches in multimodal alignment tasks, addressing both high-level semantic matching and fine-grained object-level alignment.

\usetikzlibrary{trees,positioning,shapes,shadows,arrows.meta}

\begin{figure}[ht]
    \centering
    \tikzset{
        basic/.style  = {draw, text width=2cm, align=center, font=\sffamily, rectangle},
        root/.style   = {basic, rounded corners=2pt, thin, align=center, fill=green!30},
        onode/.style = {basic, thin, rounded corners=2pt, align=center, fill=green!60, text width=2cm},
        tnode/.style = {basic, thin, fill=pink!60, text width=10em, align=left},
        xnode/.style = {basic, thin, rounded corners=2pt, align=center, fill=blue!20, text width=3.5cm},
        wnode/.style = {basic, thin, align=left, fill=pink!10!blue!80!red!10, text width=5em},
        edge from parent/.style={draw=black, edge from parent fork right}
    }

    \begin{forest}
    for tree={
        grow=east,
        growth parent anchor=west,
        parent anchor=east,
        child anchor=west,
        scale=0.7, 
        anchor=west, 
        edge path={\noexpand\path[\forestoption{edge},->, >={latex}]
             (!u.parent anchor) -- +(10pt,0pt) |-  (.child anchor)
             \forestoption{edge label};}
    }
    [Multimodal Large Language Models, basic,  l sep=7mm
        [Continual Learning, xnode,  l sep=7mm
            [{Feng et al.\, 2023\\
             Shi et al.\, 2024}, tnode]
        ]
        [Data, xnode,  l sep=7mm
            [{Bai et al.\, 2024\\
             Yang et al.\, 2024}, tnode]
        ]
        [Efficient, xnode, l sep=7mm
            [{Xu et al.\, 2024\\
             Jin et al.\, 2024\\
             Liu et al.\, 2024\\
             Yao et al.\, 2024\\
             Ma et al.\, 2024}, tnode]
        ]
        [Graphs, xnode,  l sep=7mm
            [{Jin et al.\, 2023\\
             Chen et al.\, 2024\\
             Li et al.\, 2023}, tnode]
        ]
        [RAG/info retrieval, xnode,  l sep=7mm
            [{Author Y et al.\, 2022\\
             Hu et al.\, 2024\\
             Zhao et al.\, 2023\\
             Xu et al.\, 2023\\
             Chen et al.\, 2024}, tnode]
        ]
        [Evaluation, xnode,  l sep=7mm
            [{Lu et al.\, 2024\\
             Li et al.\, 2024\\
             Chang et al.\, 2024\\
             Ge et al.\, 2024\\
             Chen et al.\, 2024}, tnode]
        ]
        [Bias, xnode,  l sep=7mm
            [{Liang et al.\, 2024\\
             Bai et al.\, 2024\\
             Gallegos et al.\, 2024\\
             Li et al.\, 2023}, tnode]
        ]
        [Agent, xnode,  l sep=7mm
            [{Wang et al.\, 2024\\
             Xi et al.\, 2023\\
             Gao et al.\, 2023\\
             Xie et al.\, 2024\\
             Xu et al.\, 2024}, tnode]
        ]
        [Security, xnode,  l sep=7mm
            [{Caffagni et al.\, 2024\\
             Zhao et al.\, 2024\\
             Wang et al.\, 2024\\
             Jin et al.\, 2024\\
             Liu et al.\, 2023\\
             Liu et al.\, 2024\\
             Shayegani et al.\, 2023\\
             Pi et al.\, 2024}, tnode]
        ]
        [Application, xnode,  l sep=7mm
            [{Minaee et al.\, 2024\\
             Cui et al.\, 2023\\
             Liu et al.\, 2023\\
             Lin et al.\, 2024\\
             Zhou et al.\, 2023\\
             Zeng et al.\, 2023\\
             Huang et al.\, 2024\\
             Gallotta et al.\, 2024\\
             Latif et al.\, 2023\\
             Tang et al.\, 2023\\
             Zhang et al.\, 2023}, tnode]
        ]
        [General, xnode,  l sep=7mm
            [{Caffagni et al.\, 2024\\
             Zhao et al.\, 2023\\
             Yin et al.\, 2023\\
             Hamadi et al.\, 2023\\
             Song et al.\, 2023\\
             Zhou et al.\, 2023\\
             Wang et al.\, 2024\\
             Qin et al.\, 2024\\
             Hadi et al.\, 2023\\
             Wu et al.\, 2023\\
             Awais et al.\, 2023\\
             Yang et al.\, 2024\\
             Sohail et al.\, 2023\\
             Kalyan et al.\, 2023}, tnode]
        ]
    ]
    \end{forest}

    \caption{Literature Survey Tree}
    \label{fig:lit_surv}
\end{figure}

\section{Taxonomy and Categorization}
\label{sec:taxonomy}

\subsection{Methodology}

This paper synthesizes findings from 58 most recent and most forefront surveys, categorized into key themes like model architectures, evaluation, applications, security, bias, and future directions. Surveys were selected based on their recency and breadth of coverage in the MLLM domain, spanning from general overviews to specific applications and challenges.

Each survey is analyzed based on:

\begin{itemize}
    \item \textbf{Technical focus:} architectures, models, datasets.
    \item \textbf{Applications:} computer vision, healthcare, robotics, etc.
    \item \textbf{Security and biases:} model safety, fairness, robustness.
    \item \textbf{Emerging trends:} future directions, new paradigms.
\end{itemize}

\subsection{Applications and Agents}

\paragraph{Law}
The survey \cite{minaee2024large} focuses on the integration of LLMs into the legal domain, where they have been used for tasks such as legal advice, document processing, and judicial assistance. The key trend here is the application of specialized LLMs, such as LawGPT and LexiLaw, which address the nuances of legal language and reasoning. One significant challenge identified is maintaining judicial independence and addressing the ethical implications of biased data in legal decision-making.

\paragraph{Autonomous Driving}
In the field of autonomous driving, MLLMs are increasingly being used to improve perception, decision-making, and human-vehicle interaction, as noted in \cite{cui2023surveymultimodallargelanguage}. A key trend is the integration of multimodal data, such as LiDAR, maps, and images, which enhance the vehicle’s ability to process complex driving environments. However, challenges related to real-time data processing and ensuring safety in diverse driving conditions remain significant.

\paragraph{Mathematics}
\cite{liu2023mathematical} explores the application of LLMs in solving mathematical tasks, such as calculation and reasoning. Techniques like Chain-of-Thought (CoT) prompting have significantly improved model performance in complex mathematical problems. However, the scarcity of high-quality datasets and the complexity of mathematical reasoning pose ongoing challenges.

\paragraph{Healthcare}
The healthcare survey \cite{lin2024multimodallearningdelivereduniversal} reviews the use of multimodal learning for tasks like image fusion, report generation, and cross-modal retrieval. A key trend is the rise of foundation models such as GPT-4 and CLIP in processing medical data. Despite advancements, these models have not yet achieved universal intelligence, and concerns related to data integration and ethical considerations remain major barriers.

\paragraph{Robotics}
The use of LLMs in robotics \cite{zeng2023large} focuses on their ability to improve perception, decision-making, and control in robotic systems. The main trend identified is the potential for LLMs to advance embodied intelligence, where robots understand and interact with the physical world. However, challenges related to real-time perception, control, and integration with existing robotic technologies persist.

\paragraph{Multilingualism}
In multilingual settings, the ability of LLMs to process multiple languages is highlighted in \cite{huang2024surveylargelanguagemodels}. There is significant progress in handling multiple languages, but low-resource languages and security issues in multilingual models remain challenges. Emerging techniques like multilingual Chain-of-Thought reasoning show promise for future development.

\paragraph{Gaming}
The survey \cite{gallotta2024large} focuses on the role of LLMs in gaming, particularly in generating dynamic Non-Player Characters (NPCs), enhancing player interaction, and even assisting in game design. The challenge of hallucinations, where LLMs generate plausible but incorrect outputs, is a major limitation in real-time gaming environments. Improving context management and memory within gaming systems is a future research priority.

\paragraph{Audio Processing}
\cite{latif2023sparks} discusses the application of LLMs in audio processing tasks, such as Automatic Speech Recognition (ASR) and music generation. The integration of multimodal data from speech, music, and environmental sounds into a single model, like SeamlessM4T, marks a significant trend. However, scalability and data diversity remain issues to address.

\paragraph{Video Understanding}
In \cite{tang2023video}, the focus is on video understanding through the use of Vid-LLMs (Video-Large Language Models), which combine video, text, and audio inputs to analyze and understand video content. While these models are promising for tasks like video summarization and captioning, challenges related to processing long videos and maintaining contextual coherence need further exploration.

\paragraph{Citation Analysis}
The use of LLMs for citation tasks is discussed in \cite{zhang2023large}, where LLMs significantly improve citation recommendation, classification, and summarization tasks. Additionally, citation data enhances LLM performance by enabling multi-hop knowledge across documents. Future research needs to address the expansion of citation networks and integration of non-academic sources.
\subsection{Evaluation and Benchmarks}
Evaluation and benchmarking of Multimodal Large Language Models (MLLMs) are crucial for understanding their performance across a diverse range of tasks and datasets. Existing evaluations can be categorized based on the models' core abilities, the datasets used, and the complexity of the tasks involved.

\subsubsection{Core Evaluation Domains}
\paragraph{Perception and Understanding}
This domain evaluates how well MLLMs interpret multimodal inputs, such as text, images, and audio, and integrate information across modalities. Benchmarks in this category include tasks like object detection, scene understanding, and feature extraction. For example, VQA datasets like VQAv2 are foundational for evaluating these abilities but are limited by biases that can inflate model performance \cite{lu2024revisiting}. 
\paragraph{Cognition and Reasoning}
Higher-level capabilities, such as logical reasoning, problem-solving, and multimodal reasoning, are captured in this domain. Tasks that require more sophisticated reasoning across modalities, like visual question answering with complex scenarios, are included in this category. This domain tests the models' deeper understanding of the relationships between the modalities \cite{li2024survey}. 

\subsubsection{Advanced Evaluation Areas}
\paragraph{Robustness and Safety}
Models need to be evaluated for their robustness, particularly when faced with adversarial prompts or out-of-distribution data. The benchmarks in this category assess how well MLLMs perform under conditions that simulate real-world challenges. Robustness is also critical in ensuring the models’ safety, especially when deployed in domains like autonomous driving or healthcare \cite{chang2024survey, niu2024large}. Furthermore, this area assesses how models manage hallucinations and mitigate biases.

\paragraph{Domain-Specific Capabilities}
This category includes evaluations focused on specialized domains, such as medical image interpretation or legal text analysis, where models must combine general multimodal abilities with domain-specific knowledge. Benchmarks like TallyQA, which test complex reasoning in specific domains, fall into this category \cite{li2024survey}.

\subsubsection{Task-Specific Benchmarks}
\paragraph{Traditional vs. Advanced Datasets}
Early datasets, such as VQAv2, have been widely used to evaluate MLLMs but exhibit limitations due to their susceptibility to language bias and lack of task complexity. In contrast, more recent datasets like TDIUC and DVQA are designed to evaluate models on fine-grained visual understanding, reasoning, and OCR tasks. These datasets provide a more rigorous assessment of the models' capabilities \cite{lu2024revisiting}.

\paragraph{Multimodal Reasoning and Interaction}
This category focuses on evaluating the interaction between modalities, which is key in tasks like multimodal dialogue or visual question answering that require reasoning across both text and images. Advanced datasets, such as VQDv1, which require complex reasoning and the identification of multiple objects in visual contexts, push models to demonstrate a deeper understanding of multimodal relationships \cite{lu2024revisiting}.

\subsubsection{Ethical and Societal Implications}
\paragraph{Fairness and Bias}
This domain addresses the importance of ensuring that MLLMs do not perpetuate or amplify societal biases. Models must be evaluated on their ability to perform equitably across demographic groups, and benchmarks in this category assess the fairness of the models \cite{li2023surveyfairness}.

\paragraph{Trustworthiness and Safety}
Ensuring trust in MLLMs is critical, especially in applications where safety is paramount. This category evaluates whether models produce harmful or misleading content and assesses their reliability in sensitive domains. It also evaluates how MLLMs handle uncertainty and avoid hallucinations \cite{gallegos2024bias, bai2024hallucination}.

In summary, evaluation and benchmarking in MLLMs have evolved from using broad, general-purpose datasets to more specialized, task-specific benchmarks that provide a deeper insight into models' true capabilities. The taxonomy proposed here synthesizes insights from existing literature, offering a structured approach to categorizing and evaluating MLLMs across multiple dimensions.

\subsubsection{Taxonomy of Benchmarking Criteria}

\textbf{Micro vs. Macro Performance:} The benchmarking process involves analyzing both micro performance (where each example is weighted equally) and macro performance (averaged across different question types). This distinction helps in understanding how well a model performs across varied tasks and datasets, providing a clearer picture of its strengths and weaknesses.

\textbf{Generalization Across Tasks:} Models are evaluated not just on their ability to perform specific tasks but also on their generalization capabilities. For example, while models like GPT-4V may excel in straightforward tasks like counting, they often face challenges in more complex reasoning tasks, revealing gaps in their overall performance.
\subsection{Efficiency and Adaptation}

The surveys on resource-efficient Multimodal Large Language Models (MLLMs) explore various approaches aimed at reducing computational costs while maintaining performance. Xu et al. \cite{xu2024survey} and Jin et al. \cite{jin2024efficientmultimodallargelanguage} emphasize the growing need for MLLMs to be more accessible, particularly in resource-constrained environments like edge computing. Both surveys provide comprehensive taxonomies, covering advancements in architectures, vision-language integration, training methods, and benchmarks that optimize MLLM efficiency. Key methods discussed include vision token compression, parameter-efficient fine-tuning, and the exploration of transformer-alternative models like Mixture of Experts (MoE) and state space models. These efforts collectively aim to balance computational efficiency with task performance, driving MLLM adoption across various practical applications.

On the other hand, Liu et al. \cite{liu2024fewshotadaptationmultimodalfoundation} address the challenge of adapting MLLMs to specific tasks with limited labeled data. The survey categorizes approaches into prompt-based, adapter-based, and external knowledge-based methods, which help these models generalize better in fine-grained domains such as medical imaging and remote sensing. Few-shot adaptation techniques, such as visual prompt tuning and adapter fine-tuning, are critical for extending the usability of large multimodal models without relying on extensive labeled datasets. Despite advancements, these surveys highlight ongoing challenges, including domain adaptation, model selection, and the integration of external knowledge. Both fields point toward a future where MLLMs are not only more efficient but also more flexible and adaptive in handling diverse, real-world tasks.

\subsection{Data-centric}

Bai et al. \cite{bai2024surveymultimodallargelanguage} presents a comprehensive survey on the role of data in the development of Multimodal Large Language Models (MLLMs). The paper emphasizes the importance of the quality, diversity, and volume of multimodal data, which includes text, images, audio, and video, in training MLLMs effectively. It identifies significant challenges in multimodal data collection, such as data sparsity and noise, and explores potential solutions like synthetic data generation and active learning to mitigate these issues. The authors advocate for a more data-centric approach, where the refinement and curation of data take precedence, ultimately improving model performance and advancing MLLM capabilities in an increasingly complex landscape.

Yang et al. \cite{yang2024if} investigates the intersection of large language models (LLMs) and code, framing code as a vital tool that enhances LLMs' capacity to operate as intelligent agents. The paper discusses how code empowers LLMs in tasks such as automation, code generation, and software development. It also tackles key challenges around code correctness, efficiency, and security, which are essential when deploying LLMs in practical applications. By analyzing the integration of code with LLMs, the authors showcase the potential of these models to evolve into autonomous agents capable of managing complex tasks across various domains, thereby broadening the scope and impact of LLMs in AI-driven innovations.

\subsection{Contunual Learning}

Feng et al. \cite{feng2023trends} provides a thorough overview of how Large Language Models (LLMs) are being integrated with external knowledge to enhance their capabilities. The survey categorizes the integration methods into two main approaches: knowledge editing and retrieval augmentation. Knowledge editing involves modifying the input or the model itself to update the outdated or incorrect information, while retrieval augmentation fetches external information during inference without altering the model’s core parameters. The authors present a taxonomy covering these methods, benchmarks for evaluation, and applications such as LangChain and ChatDoctor, which leverage these strategies to address domain-specific challenges. Additionally, the paper explores the handling of knowledge conflicts and suggests future research directions for improving LLM performance in complex, real-world tasks through better integration of multi-source knowledge.

Shi et al. \cite{shi2024continual} offer an extensive survey on the continual learning (CL) of Large Language Models (LLMs), addressing the critical challenges and methodologies in this field. presents an extensive overview of the challenges and techniques related to the continual learning (CL) of Large Language Models (LLMs). The authors focus on two primary directions of continuity: vertical continual learning (adapting models from general to specific domains) and horizontal continual learning (adapting models over time across various domains). They discuss the problems of "catastrophic forgetting," where models lose knowledge of previous tasks, and the complexity of continually updating models to maintain performance on both old and new tasks. The survey outlines key CL methods, including continual pre-training, domain-adaptive pre-training, and continual fine-tuning. It also evaluates various CL techniques, such as replay-based, regularization-based, and architecture-based methods, to mitigate forgetting and ensure knowledge retention. The authors call for more research into evaluation benchmarks and methodologies to counter forgetting and support knowledge transfer in LLMs, making this an underexplored yet crucial area of machine learning research
\subsection{Evaluation Benchmarks}

Lu et al. \cite{lu2024revisiting} tackle the key challenges in evaluating multimodal large language models (MLLMs), focusing on the unique complexities these models present. They propose a taxonomy of evaluation metrics designed specifically for multimodal tasks, such as cross-modal retrieval, caption generation, and visual question answering (VQA). The authors highlight how current evaluation frameworks often fail to capture the nuances of multimodal interactions, suggesting the need for more tailored metrics to address these limitations.

Li and Lu \cite{li2024surveybenchmarksmultimodallarge} provide a comprehensive overview of benchmarking datasets and performance metrics in MLLMs. They argue that the absence of standardized protocols across different modalities hinders the fair comparison of models. Their work calls for the establishment of consistent evaluation frameworks that ensure reliable and equitable performance assessment across varied multimodal tasks.

Chang et al. \cite{chang2024survey} examine the evaluation methodologies for large language models (LLMs), emphasizing the importance of moving beyond task-specific performance metrics like knowledge reasoning. They point out that many existing benchmarks, such as HELM and BIG-Bench, overlook critical issues such as hallucinations, fairness, and societal implications. The authors advocate for more holistic evaluation practices that take into account not only the technical capabilities of LLMs but also their trustworthiness, robustness, and ethical alignment in real-world applications. Their work underscores the need to assess both the technical and societal impacts of these models to ensure responsible AI deployment.
\subsection{Agents and Autonomous Systems}

Xi et al. \cite{xi2023rise}, Wang et al. \cite{wang23surveyonllmautoagents} and a few more papers\cite{xie24lma, gao2023large, xu2024survey} provide comprehensive surveys on the development of LLM-based autonomous agents, highlighting the key modules and frameworks that form the foundation of these systems. At their core, autonomous agents leveraging large language models (LLMs) rely on four essential components: perception, memory, planning, and action. These modules work in synergy to enable agents to perceive their environment, recall previous interactions, and plan and execute actions in real-time. As Xi et al. \cite{xi2023rise} describe, this architecture allows agents to be highly adaptable across a variety of domains, from digital assistants to autonomous vehicles. Wang et al. \cite{wang23surveyonllmautoagents} further emphasize the importance of expanding the perception-action loop by incorporating multimodal inputs, ensuring agents can effectively handle complex real-world scenarios, such as those found in industrial automation and gaming.

Despite these advancements, LLM-based agents face several critical challenges. Both surveys identify knowledge boundaries as a significant issue, where agents are constrained when operating in specialized or underexplored domains. Prompt robustness is another challenge, as even minor changes in prompts can lead to unpredictable or erroneous behavior, including hallucinations, where agents generate false information. Shi et al. \cite{shi2024continual} further explore the challenge of catastrophic forgetting, where agents fail to retain knowledge from previous tasks after being updated with new information. Addressing these challenges requires ongoing improvements in how agents interact with external tools, refine prompt structures, and manage memory to prevent knowledge decay and ensure consistent behavior.

LLM-based agents have demonstrated significant versatility across a wide range of applications. Xi et al. \cite{xi2023rise} discuss various applications involving single agents, multi-agent systems, and human-agent interactions. Wang et al. \cite{wang23surveyonllmautoagents} categorize applications into three key areas: social sciences, natural sciences, and engineering, illustrating the transformative potential of LLM-based agents in empowering these fields. Xie et al. \cite{xie24lma} further categorize agents by their application scenarios, highlighting areas such as robotics and embodied AI, where agents make real-time decisions based on multimodal inputs like images, sensor data, and text. Additionally, as noted by Xi et al. \cite{xi2023rise} and Xie et al. \cite{xie24lma}, LLM-based agents are becoming increasingly prominent in scientific research, automating tasks such as experiment design, planning, and execution.

To ensure the robustness and reliability of LLM-based agents, future research must focus on overcoming the current limitations. Xie et al. \cite{xie24lma} and Shi et al. \cite{shi2024continual} emphasize the need to develop more robust multimodal perception systems, improve LLM inference efficiency, and establish ethical frameworks to guide agent decision-making. Enhancing memory integration and refining prompt design will also be critical to preventing issues like hallucinations and ensuring agents can effectively transfer knowledge across tasks without degradation.
\subsection{MLLMs in Graph Learning}

Multimodal large language models (MLLMs) have been increasingly applied to graph learning tasks, outperforming traditional graph neural networks (GNNs). By incorporating textual attributes and other modalities, MLLMs enhance the representational power of GNNs, enabling improved performance in classification, prediction, and reasoning tasks. Jin et al. proposed a taxonomy categorizing the integration of MLLMs with graphs into enhancers, predictors, and alignment components, highlighting their utility in various graph tasks. Similarly, Chen et al. and Li et al. discuss the integration of LLMs with knowledge graphs, illustrating the benefits of combining graph structures with multimodal data for broader applications~\cite{jin2023large, chen2024knowledge, li2023survey}.

\subsection{Retrieval-Augmented Generation (RAG) in MLLM}

Multimodal Large Language Models (MLLMs) have demonstrated remarkable capabilities in generating and understanding content across various modalities, such as text, images, and audio. However, their reliance on static training data limits their ability to provide accurate, up-to-date responses, especially in rapidly changing contexts. Retrieval-Augmented Generation (RAG) addresses this issue by dynamically retrieving relevant external information before the generation process. By incorporating real-time and contextually accurate information, RAG enhances both the factuality and robustness of MLLM outputs. In multimodal tasks, RAG not only retrieves textual data but also incorporates multimodal data sources like images and videos, significantly improving the knowledge richness and generation quality of MLLMs \cite{hu2024rag, chen2024mllmstrongrerankeradvancing}.

Hu et al. \cite{hu2024rag} provide a comprehensive survey of RAG's application in natural language processing, highlighting how retrieving external knowledge during generation effectively mitigates long-tail knowledge gaps and hallucination issues. Zhao et al. \cite{zhao2023retrievingmultimodalinformationaugmented} further explore how multimodal information retrieval augments generation by improving diversity and robustness. By leveraging information from different modalities, RAG empowers MLLMs to generate more accurate and contextually grounded outputs, especially in tasks requiring cross-modal reasoning such as visual question answering and complex dialogue generation \cite{xu2023large, chen2024mllmstrongrerankeradvancing}.

\section{Major Themes Emerging from Surveys}

\subsection{MLLM Architectures}
A recurring topic across most surveys is the architecture of MLLMs, where Transformer-based models dominate. Innovations like CLIP, DALL-E, and Flamingo exemplify the progress in aligning text and visual data. Surveys compare early fusion (integrating modalities early in the model pipeline) and late fusion strategies (processing modalities separately before combining outputs).

\subsection{Datasets and Training}
The field is characterized by massive, multimodal datasets such as MS-COCO, Visual Genome, and custom-curated sets like LAION. Pretraining on large-scale datasets remains a core strategy, with some surveys, like ``Large-scale Multi-Modal Pre-trained Models'', etc.\cite{wang2024largescalemultimodalpretrainedmodels}, offering a comprehensive taxonomy of pretraining methodologies.

\subsection{Evaluation and Metrics}
Evaluation challenges are a key theme, where traditional language or vision metrics fall short. Cross-modal retrieval, image captioning, and visual question answering (VQA) are popular benchmarks, but new methods to evaluate multimodal coherence and reasoning are discussed in evaluation-focused surveys like ``A Survey on Evaluation of Large Language Models'', etc.\cite{chang2024survey, ge2024mllmbenchevaluatingmultimodalllms, chen2024mllmasajudgeassessingmultimodalllmasajudge}.

\subsubsection{Security: Adversarial Attacks}

Adversarial attacks have been a critical concern for MLLMs. Caffagni et al.\cite{caffagni2024revolutionmultimodallargelanguage} highlight that MLLMs are particularly susceptible to \textbf{adversarial attacks} that exploit weaknesses in visual inputs. By introducing imperceptible noise into images, attackers can manipulate the vision encoder, leading to incorrect or even harmful text generation. Wang et al.\cite{wang2024llms} elaborate on \textbf{jailbreaking attacks}, where adversaries bypass the model's safety alignment mechanisms. Techniques like \textbf{prompt injection} can disrupt the model’s \textbf{chain-of-thought reasoning}, leading to the generation of dangerous or inappropriate content. Especially in \textbf{white-box attacks}, attackers utilize the model’s \textbf{gradient information} to craft adversarial examples that precisely control the model’s outputs.\cite{pi2024mllmprotectorensuringmllmssafety}

Zhao et al.\cite{zhao2024survey} emphasize the complexity of \textbf{multimodal adversarial attacks}, where both image and text inputs are manipulated. These attacks exploit the model's difficulty in handling cross-modal perturbations, making it challenging to detect malicious inputs. To counter these attacks, researchers suggest improving cross-modal alignment algorithms and developing new defense mechanisms to enhance robustness across modalities.

Another vulnerability arises in the \textbf{cross-modal alignment} process, which connects the features extracted from images with those from textual data. As Shayegani et al.\cite{shayegani2023survey} explain, attackers can exploit this process to disrupt the model’s understanding of multimodal data, compromising its predictions. This highlights a need for robust alignment mechanisms that can defend against such manipulations.

\subsubsection{Bias: Hallucinations and Data Bias}

Liu et al.\cite{liu2024surveyhallucinationlargevisionlanguage} describe the \textbf{hallucination phenomenon} in Multimodal Large Language Models (MLLMs), where the generated outputs deviate from the visual input, leading to the fabrication of objects or misrepresentation of relationships in an image. This issue is often a byproduct of \textbf{data bias} in the training sets, as well as the limitations of vision encoders in accurately grounding images.

Hallucinations can be traced to several causes, ranging from the models' reliance on noisy or incomplete training data to the inherent limitations of cross-modal alignment mechanisms\cite{bai2024hallucination}. For instance, models may incorporate statistically frequent objects that are not present in the specific image or confuse the relationships between depicted elements. This happens because MLLMs may memorize biases present in training datasets, such as common object pairings or overly simplistic image-text associations. As a result, hallucinations not only mislead users but also exacerbate \textbf{existing biases} in the models' outputs, leading to incorrect assumptions about objects' existence, attributes, or relationships\cite{gallegos2024bias}.

Addressing these hallucinations requires improving the \textbf{modality alignment mechanisms} to better synchronize textual and visual data. This involves enhancing the capabilities of vision encoders to more accurately represent fine-grained details of the images and ensuring that the language models respect the constraints imposed by visual context. Bai et al.\cite{bai2024hallucination} advocate for the use of specialized evaluation benchmarks and techniques, such as leveraging better-aligned datasets or employing post-processing corrections. Furthermore, strategies like counterfactual data augmentation and the introduction of diverse visual instructions can help models generalize better to unseen scenarios, thus reducing the frequency of hallucinations and aligning model predictions with true visual inputs\cite{liang2024internal}.

\subsubsection{Fairness: Adversarial Training and Human Feedback}

To address both security and fairness, several defense strategies have been proposed. Shayegani et al.\cite{shayegani2023survey} suggest employing \textbf{adversarial training} and \textbf{data augmentation} as methods to strengthen the model’s robustness. By introducing adversarial examples during the training phase, models can learn to identify and resist adversarial inputs. Furthermore, \textbf{safety steering vectors} and \textbf{multimodal input validation} can dynamically detect and correct potentially biased or adversarial outputs during inference.

Liu et al.\cite{liu2023trustworthy} explore techniques to improve fairness by leveraging \textbf{Reinforcement Learning from Human Feedback (RLHF)}. This method allows models to adjust their outputs based on user feedback, ensuring that the generated text aligns with societal values and ethical standards. The approach ensures that models maintain fairness by incorporating diverse viewpoints and mitigating biases present in training data.

\subsubsection{Defense Against Jailbreaking Attacks}

Jin et al.\cite{jin2024jailbreakzoo} classify different types of \textbf{jailbreaking attacks} and propose defense strategies such as \textbf{prompt tuning} and \textbf{gradient-based defenses}. These approaches focus on adjusting the input prompts and strengthening alignment algorithms to reduce the success rate of jailbreaking attacks, ensuring that the model remains secure and aligned with ethical guidelines.

In conclusion, the security, bias, and fairness challenges in MLLMs primarily revolve around defending against adversarial attacks, addressing training data biases, and ensuring fair handling of multimodal inputs. Future research must focus on optimizing vision encoders and cross-modal alignment mechanisms to improve the robustness and fairness of MLLMs.
\section{Emerging Trends and Gaps}

\label{sec:trends_gaps}

\subsection{Current Trends}

As Multimodal Large Language Models continues to advance, several key trends have emerged that reflect both the growing capabilities and the challenges associated with these models. Recent surveys provide valuable insights into how MLLMs are evolving and highlight significant areas of focus that are shaping the current landscape of research and application. Below, we outline the most discussed and impactful trends observed in the literature.

\textbf{Increased Integration of Multimodality}

One of the most prominent trends in the surveyed literature is the enhanced integration of multiple modalities—such as text, images, and audio—within MLLMs. Surveys like ``The (R)Evolution of Multimodal Large Language Models'', etc.\cite{caffagni2024r, hamadi2023large, song2023bridge, zhou2023comprehensive} emphasize the shift from unimodal to multimodal systems as a transformative leap, enabling models to more closely mimic human perception. This trend is echoed across multiple papers, highlighting the field's focus on achieving a more holistic understanding of information by combining different types of data inputs.

\textbf{Applications in Diverse Domains}

A recurring theme in recent surveys is the expansion of MLLM applications into various domains, particularly those requiring complex, multimodal understanding, such as autonomous agents and medical diagnostics. For instance, the survey ``A Survey on Multimodal Large Language Models''\cite{yin2023survey} traces the significant impact of MLLMs across different industries, suggesting a growing trend toward domain-specific adaptations of these models. This trend points to the increasing specialization of MLLMs to meet the demands of specific fields.

\textbf{Focus on Evaluation Metrics and Benchmarking}

As MLLMs become more prevalent, the need for robust evaluation metrics and benchmarking has become increasingly critical. The surveyed literature frequently discusses the development of new benchmarks designed to assess the performance of MLLMs across different modalities. For example, ``A Survey of Large Language Models''\cite{zhao2023surveylargelanguagemodels} provides an in-depth analysis of existing evaluation methodologies, indicating a trend towards more comprehensive and standardized performance assessments.

\textbf{Ethical and Security Considerations}

Another notable trend is the growing concern over the ethical implications and security risks associated with MLLMs. Recent surveys reflect an increased focus on the responsible development and deployment of these models. Ethical concerns, such as bias in multimodal data processing and the potential for misuse in generating deceptive content, are frequently discussed, highlighting the importance of addressing these issues as the technology advances.

\textbf{Efficiency and Optimization}

Efficiency in training and deploying MLLMs is an emerging trend, particularly as models become larger and more complex. The surveyed literature suggests a focus on optimizing the computational resources required for MLLMs, making them more accessible and scalable. This trend is crucial for the broader adoption of these models in both research and industry.

These trends underscore the rapid evolution of MLLMs and the diverse challenges and opportunities they present. The surveyed literature not only identifies these key areas but also sets the stage for future research directions, which are discussed in the subsequent sections. \cite{yao2024minicpmvgpt4vlevelmllm, ma2024eemllmdataefficientcomputeefficientmultimodal}

\subsection{Research Gaps}

Certain areas of MLLMs remain underexplored or lack comprehensive analysis. Identifying these research gaps guides future studies to ensure a balanced development in the field. Below, we highlight key areas that needs further investigation.

\textbf{Integration of Lesser-Known Modalities}

While most surveys focus on the integration of text, images, and, to some extent, audio, there is a noticeable gap in exploring the potential of other modalities, such as haptic feedback, olfactory data, and advanced sensory inputs. The survey like "The (R)Evolution of Multimodal Large Language Models", etc.\cite{caffagni2024r, hadi2023large, 10386743, awais2023foundational, yang2024harnessing} acknowledges the importance of holistic integration but primarily emphasizes traditional modalities. Future research could expand on how these lesser-known modalities can be incorporated into MLLMs to create even more comprehensive models.

\textbf{Longitudinal Studies on Model Performance}

Current surveys tend to focus on the performance of MLLMs at a specific point in time, often lacking longitudinal studies that track the evolution of model capabilities and limitations over extended periods. For instance, "A Survey of Large Language Models"\cite{zhao2023surveylargelanguagemodels, sohail2023future, kalyan2023survey, carolan2024review} provides a snapshot of the state of MLLMs but does not address how these models might evolve with new data, architectures, or training techniques. Longitudinal studies could provide valuable insights into the long-term viability and scalability of MLLMs.

\textbf{Cross-Domain Applications and Transfer Learning}

There is limited coverage on the application of MLLMs across vastly different domains and the effectiveness of transfer learning in these contexts. Surveys such as "Large-scale Multi-Modal Pre-trained Models: A Survey"\cite{wang2024largescalemultimodalpretrainedmodels} touch upon domain-specific applications but do not fully explore how models trained in one domain perform when transferred to another. This is a critical area for research, especially in understanding the generalization capabilities of MLLMs across diverse fields.

\textbf{Ethical Implications in Non-Textual Modalities}

While ethical concerns in textual data have been widely discussed, there is a research gap in exploring the ethical implications of non-textual modalities, particularly in images and video. The survey "Multimodal Large Language Models: A Survey"\cite{10386743} addresses general ethical concerns but lacks a deep dive into how these issues manifest in non-textual data. Future research should focus on understanding the ethical challenges unique to each modality, including biases, privacy concerns, and the potential for misuse.

\textbf{Impact of Multilingualism on Multimodality}

The intersection of multilingualism and multimodality is another underexplored area. Although "Multilingual Large Language Model: A Survey of Surveys"\cite{qin2024multilinguallargelanguagemodel} addresses multilingual capabilities, it does not fully explore how these capabilities interact with multimodal inputs. Research in this area could lead to more inclusive MLLMs that better serve global populations by effectively integrating multilingual and multimodal data.

\section{Conclusion}
\label{sec:conclusion}

\subsection{Summary of Insights}
In our survey of surveys on Multimodal Large Language Models (MLLMs), several key insights have emerged. First, the multimodal capabilities of MLLMs have significantly enhanced performance across various tasks, particularly in combining and understanding data from multiple sources such as text, images, and videos. Second, as the size of models and the amount of training data increases, the intelligence of the models in perception and reasoning improves, but at the cost of increasing computational resource demands. Additionally, we observed challenges related to robustness, interpretability, and fairness of the models in practical applications. Finally, MLLMs are expanding their potential applications across numerous industries, driven by rapid technological advancements.

\subsection{Future Directions}
Future surveys can be improved and expanded in the following ways to better capture emerging trends in the rapidly evolving field of MLLMs:

\begin{itemize}
    \item \textbf{Emerging Areas}: With the development of generative AI and self-supervised learning, future surveys should focus on new trends emerging from the integration of these technologies with MLLMs.
    \item \textbf{Data Diversity and Challenges}: There should be a deeper discussion of the challenges posed by the diversity and complexity of multimodal data, particularly regarding the construction, annotation, and management of large-scale datasets.
    \item \textbf{Model Evaluation and Standardization}: Future surveys should systematically analyze evaluation standards, including performance metrics across different tasks and domains, and evaluate the robustness and fairness of the models.
    \item \textbf{Real-World Applications and Ethics}: A more thorough examination of real-world applications of MLLMs is needed, including issues of privacy, security, and ethical considerations, with recommendations on how to balance innovation and risks in various application scenarios.
    \item \textbf{Optimization of Computational Resources}: There should be further exploration of efficient use of computational resources and model compression techniques, providing guidance for MLLM applications in resource-constrained environments.
\end{itemize}

\bibliographystyle{unsrt}  
\bibliography{references}

@article{lu2024revisiting,
  title={Revisiting Multi-Modal LLM Evaluation},
  author={Lu, Jian and Srivastava, Shikhar and Chen, Junyu and Shrestha, Robik and Acharya, Manoj and Kafle, Kushal and Kanan, Christopher},
  journal={arXiv preprint arXiv:2408.05334},
  year={2024}
}

@misc{li2024survey,
      title={A Survey on Benchmarks of Multimodal Large Language Models}, 
      author={Jian Li and Weiheng Lu},
      year={2024},
      eprint={2408.08632},
      archivePrefix={arXiv},
      primaryClass={cs.CL},
      url={https://arxiv.org/abs/2408.08632}, 
}

@article{chang2024survey,
  title={A survey on evaluation of large language models},
  author={Chang, Yupeng and Wang, Xu and Wang, Jindong and Wu, Yuan and Yang, Linyi and Zhu, Kaijie and Chen, Hao and Yi, Xiaoyuan and Wang, Cunxiang and Wang, Yidong and others},
  journal={ACM Transactions on Intelligent Systems and Technology},
  volume={15},
  number={3},
  pages={1--45},
  year={2024},
  publisher={ACM New York, NY}
}

@misc{caffagni2024revolutionmultimodallargelanguage,
      title={The Revolution of Multimodal Large Language Models: A Survey}, 
      author={Davide Caffagni and Federico Cocchi and Luca Barsellotti and Nicholas Moratelli and Sara Sarto and Lorenzo Baraldi and Lorenzo Baraldi and Marcella Cornia and Rita Cucchiara},
      year={2024},
      eprint={2402.12451},
      archivePrefix={arXiv},
      primaryClass={cs.CV},
      url={https://arxiv.org/abs/2402.12451}, 
}

@inproceedings{zhao2024survey,
  title={A Survey on Safe Multi-Modal Learning Systems},
  author={Zhao, Tianyi and Zhang, Liangliang and Ma, Yao and Cheng, Lu},
  booktitle={Proceedings of the 30th ACM SIGKDD Conference on Knowledge Discovery and Data Mining},
  pages={6655--6665},
  year={2024}
}

@article{wang2024llms,
  title={From LLMs to MLLMs: Exploring the Landscape of Multimodal Jailbreaking},
  author={Wang, Siyuan and Long, Zhuohan and Fan, Zhihao and Wei, Zhongyu},
  journal={arXiv preprint arXiv:2406.14859},
  year={2024}
}

@article{jin2024jailbreakzoo,
  title={JailbreakZoo: Survey, Landscapes, and Horizons in Jailbreaking Large Language and Vision-Language Models},
  author={Jin, Haibo and Hu, Leyang and Li, Xinuo and Zhang, Peiyan and Chen, Chonghan and Zhuang, Jun and Wang, Haohan},
  journal={arXiv preprint arXiv:2407.01599},
  year={2024}
}

@article{liu2023trustworthy,
  title={Trustworthy LLMs: A survey and guideline for evaluating large language models' alignment},
  author={Liu, Yang and Yao, Yuanshun and Ton, Jean-Francois and Zhang, Xiaoying and Guo, Ruocheng and Cheng, Hao and Klochkov, Yegor and Taufiq, Muhammad Faaiz and Li, Hang},
  journal={arXiv preprint arXiv:2308.05374},
  year={2023}
}

@misc{liu2024surveyhallucinationlargevisionlanguage,
      title={A Survey on Hallucination in Large Vision-Language Models}, 
      author={Hanchao Liu and Wenyuan Xue and Yifei Chen and Dapeng Chen and Xiutian Zhao and Ke Wang and Liping Hou and Rongjun Li and Wei Peng},
      year={2024},
      eprint={2402.00253},
      archivePrefix={arXiv},
      primaryClass={cs.CV},
      url={https://arxiv.org/abs/2402.00253}, 
}

@article{shayegani2023survey,
  title={Survey of vulnerabilities in large language models revealed by adversarial attacks},
  author={Shayegani, Erfan and Mamun, Md Abdullah Al and Fu, Yu and Zaree, Pedram and Dong, Yue and Abu-Ghazaleh, Nael},
  journal={arXiv preprint arXiv:2310.10844},
  year={2023}
}

@misc{li2024surveybenchmarksmultimodallarge,
      title={A Survey on Benchmarks of Multimodal Large Language Models}, 
      author={Jian Li and Weiheng Lu},
      year={2024},
      eprint={2408.08632},
      archivePrefix={arXiv},
      primaryClass={cs.CL},
      url={https://arxiv.org/abs/2408.08632}, 
}

@article{caffagni2024r,
  title={The (r) evolution of multimodal large language models: A survey},
  author={Caffagni, Davide and Cocchi, Federico and Barsellotti, Luca and Moratelli, Nicholas and Sarto, Sara and Baraldi, Lorenzo and Cornia, Marcella and Cucchiara, Rita},
  journal={arXiv preprint arXiv:2402.12451},
  year={2024}
}

@misc{zhao2023surveylargelanguagemodels,
      title={A Survey of Large Language Models}, 
      author={Wayne Xin Zhao and Kun Zhou and Junyi Li and Tianyi Tang and Xiaolei Wang and Yupeng Hou and Yingqian Min and Beichen Zhang and Junjie Zhang and Zican Dong and Yifan Du and Chen Yang and Yushuo Chen and Zhipeng Chen and Jinhao Jiang and Ruiyang Ren and Yifan Li and Xinyu Tang and Zikang Liu and Peiyu Liu and Jian-Yun Nie and Ji-Rong Wen},
      year={2023},
      eprint={2303.18223},
      archivePrefix={arXiv},
      primaryClass={cs.CL},
      url={https://arxiv.org/abs/2303.18223}, 
}

@article{yin2023survey,
  title={A survey on multimodal large language models},
  author={Yin, Shukang and Fu, Chaoyou and Zhao, Sirui and Li, Ke and Sun, Xing and Xu, Tong and Chen, Enhong},
  journal={arXiv preprint arXiv:2306.13549},
  year={2023}
}

@article{hamadi2023large,
  title={Large Language Models Meet Computer Vision: A Brief Survey},
  author={Hamadi, Raby},
  journal={arXiv preprint arXiv:2311.16673},
  year={2023}
}

@article{song2023bridge,
  title={How to bridge the gap between modalities: A comprehensive survey on multimodal large language model},
  author={Song, Shezheng and Li, Xiaopeng and Li, Shasha},
  journal={arXiv preprint arXiv:2311.07594},
  year={2023}
}

@article{zhou2023comprehensive,
  title={A comprehensive survey on pretrained foundation models: A history from bert to chatgpt},
  author={Zhou, Ce and Li, Qian and Li, Chen and Yu, Jun and Liu, Yixin and Wang, Guangjing and Zhang, Kai and Ji, Cheng and Yan, Qiben and He, Lifang and others},
  journal={arXiv preprint arXiv:2302.09419},
  year={2023}
}

@misc{wang2024largescalemultimodalpretrainedmodels,
      title={Large-scale Multi-Modal Pre-trained Models: A Comprehensive Survey}, 
      author={Xiao Wang and Guangyao Chen and Guangwu Qian and Pengcheng Gao and Xiao-Yong Wei and Yaowei Wang and Yonghong Tian and Wen Gao},
      year={2024},
      eprint={2302.10035},
      archivePrefix={arXiv},
      primaryClass={cs.CV},
      url={https://arxiv.org/abs/2302.10035}, 
}

@misc{qin2024multilinguallargelanguagemodel,title={Multilingual Large Language Model: A Survey of Resources, Taxonomy and Frontiers}, author={Libo Qin and Qiguang Chen and Yuhang Zhou and Zhi Chen and Yinghui Li and Lizi Liao and Min Li and Wanxiang Che and Philip S. Yu},year={2024},eprint={2404.04925},archivePrefix={arXiv},primaryClass={cs.CL},url={https://arxiv.org/abs/2404.04925}
}

@article{hadi2023large,
  title={Large language models: a comprehensive survey of its applications, challenges, limitations, and future prospects},
  author={Hadi, Muhammad Usman and Qureshi, Rizwan and Shah, Abbas and Irfan, Muhammad and Zafar, Anas and Shaikh, Muhammad Bilal and Akhtar, Naveed and Wu, Jia and Mirjalili, Seyedali and others},
  journal={Authorea Preprints},
  year={2023},
  publisher={Authorea}
}

@INPROCEEDINGS {10386743,
author = {J. Wu and W. Gan and Z. Chen and S. Wan and P. S. Yu},
booktitle = {2023 IEEE International Conference on Big Data (BigData)},
title = {Multimodal Large Language Models: A Survey},
year = {2023},
volume = {},
issn = {},
pages = {2247-2256},
doi = {10.1109/BigData59044.2023.10386743},
url = {https://doi.ieeecomputersociety.org/10.1109/BigData59044.2023.10386743},
publisher = {IEEE Computer Society},
address = {Los Alamitos, CA, USA},
month = {dec}
}

@article{awais2023foundational,
  title={Foundational models defining a new era in vision: A survey and outlook},
  author={Awais, Muhammad and Naseer, Muzammal and Khan, Salman and Anwer, Rao Muhammad and Cholakkal, Hisham and Shah, Mubarak and Yang, Ming-Hsuan and Khan, Fahad Shahbaz},
  journal={arXiv preprint arXiv:2307.13721},
  year={2023}
}

@article{yang2024harnessing,
  title={Harnessing the power of llms in practice: A survey on chatgpt and beyond},
  author={Yang, Jingfeng and Jin, Hongye and Tang, Ruixiang and Han, Xiaotian and Feng, Qizhang and Jiang, Haoming and Zhong, Shaochen and Yin, Bing and Hu, Xia},
  journal={ACM Transactions on Knowledge Discovery from Data},
  volume={18},
  number={6},
  pages={1--32},
  year={2024},
  publisher={ACM New York, NY}
}

@article{sohail2023future,
  title={The future of gpt: A taxonomy of existing chatgpt research, current challenges, and possible future directions},
  author={Sohail, Shahab Saquib and Farhat, Faiza and Himeur, Yassine and Nadeem, Mohammad and Madsen, Dag {\O}ivind and Singh, Yashbir and Atalla, Shadi and Mansoor, Wathiq},
  journal={Current Challenges, and Possible Future Directions (April 8, 2023)},
  year={2023}
}

@article{kalyan2023survey,
  title={A survey of GPT-3 family large language models including ChatGPT and GPT-4},
  author={Kalyan, Katikapalli Subramanyam},
  journal={Natural Language Processing Journal},
  pages={100048},
  year={2023},
  publisher={Elsevier}
}

@article{liang2024internal,
  title={Internal consistency and self-feedback in large language models: A survey},
  author={Liang, Xun and Song, Shichao and Zheng, Zifan and Wang, Hanyu and Yu, Qingchen and Li, Xunkai and Li, Rong-Hua and Xiong, Feiyu and Li, Zhiyu},
  journal={arXiv preprint arXiv:2407.14507},
  year={2024}
}

@article{bai2024hallucination,
  title={Hallucination of multimodal large language models: A survey},
  author={Bai, Zechen and Wang, Pichao and Xiao, Tianjun and He, Tong and Han, Zongbo and Zhang, Zheng and Shou, Mike Zheng},
  journal={arXiv preprint arXiv:2404.18930},
  year={2024}
}

@article{gallegos2024bias,
  title={Bias and fairness in large language models: A survey},
  author={Gallegos, Isabel O and Rossi, Ryan A and Barrow, Joe and Tanjim, Md Mehrab and Kim, Sungchul and Dernoncourt, Franck and Yu, Tong and Zhang, Ruiyi and Ahmed, Nesreen K},
  journal={Computational Linguistics},
  pages={1--79},
  year={2024},
  publisher={MIT Press 255 Main Street, 9th Floor, Cambridge, Massachusetts 02142, USA~…}
}

@article{li2023surveyfairness,
  title={A survey on fairness in large language models},
  author={Li, Yingji and Du, Mengnan and Song, Rui and Wang, Xin and Wang, Ying},
  journal={arXiv preprint arXiv:2308.10149},
  year={2023}
}

@article{xi2023rise,
  title={The rise and potential of large language model based agents: A survey},
  author={Xi, Zhiheng and Chen, Wenxiang and Guo, Xin and He, Wei and Ding, Yiwen and Hong, Boyang and Zhang, Ming and Wang, Junzhe and Jin, Senjie and Zhou, Enyu and others},
  journal={arXiv preprint arXiv:2309.07864},
  year={2023}
}

@article{gao2023large,
  title={Large language models empowered agent-based modeling and simulation: A survey and perspectives},
  author={Gao, Chen and Lan, Xiaochong and Li, Nian and Yuan, Yuan and Ding, Jingtao and Zhou, Zhilun and Xu, Fengli and Li, Yong},
  journal={arXiv preprint arXiv:2312.11970},
  year={2023}
}

@article{xu2024survey,
  title={A survey on game playing agents and large models: Methods, applications, and challenges},
  author={Xu, Xinrun and Wang, Yuxin and Xu, Chaoyi and Ding, Ziluo and Jiang, Jiechuan and Ding, Zhiming and Karlsson, B{\o}rje F},
  journal={arXiv preprint arXiv:2403.10249},
  year={2024}
}

@article{minaee2024large,
  title={Large language models: A survey},
  author={Minaee, Shervin and Mikolov, Tomas and Nikzad, Narjes and Chenaghlu, Meysam and Socher, Richard and Amatriain, Xavier and Gao, Jianfeng},
  journal={arXiv preprint arXiv:2402.06196},
  year={2024}
}

@misc{cui2023surveymultimodallargelanguage,
      title={A Survey on Multimodal Large Language Models for Autonomous Driving}, 
      author={Can Cui and Yunsheng Ma and Xu Cao and Wenqian Ye and Yang Zhou and Kaizhao Liang and Jintai Chen and Juanwu Lu and Zichong Yang and Kuei-Da Liao and Tianren Gao and Erlong Li and Kun Tang and Zhipeng Cao and Tong Zhou and Ao Liu and Xinrui Yan and Shuqi Mei and Jianguo Cao and Ziran Wang and Chao Zheng},
      year={2023},
      eprint={2311.12320},
      archivePrefix={arXiv},
      primaryClass={cs.AI},
      url={https://arxiv.org/abs/2311.12320}, 
}

@article{liu2023mathematical,
  title={Mathematical language models: A survey},
  author={Liu, Wentao and Hu, Hanglei and Zhou, Jie and Ding, Yuyang and Li, Junsong and Zeng, Jiayi and He, Mengliang and Chen, Qin and Jiang, Bo and Zhou, Aimin and others},
  journal={arXiv preprint arXiv:2312.07622},
  year={2023}
}

@misc{lin2024multimodallearningdelivereduniversal,
      title={Has Multimodal Learning Delivered Universal Intelligence in Healthcare? A Comprehensive Survey}, 
      author={Qika Lin and Yifan Zhu and Xin Mei and Ling Huang and Jingying Ma and Kai He and Zhen Peng and Erik Cambria and Mengling Feng},
      year={2024},
      eprint={2408.12880},
      archivePrefix={arXiv},
      primaryClass={cs.AI},
      url={https://arxiv.org/abs/2408.12880}, 
}

@article{zeng2023large,
  title={Large language models for robotics: A survey},
  author={Zeng, Fanlong and Gan, Wensheng and Wang, Yongheng and Liu, Ning and Yu, Philip S},
  journal={arXiv preprint arXiv:2311.07226},
  year={2023}
}

@misc{huang2024surveylargelanguagemodels,title={A Survey on Large Language Models with Multilingualism: Recent Advances and New Frontiers}, author={Kaiyu Huang and Fengran Mo and Hongliang Li and You Li and Yuanchi Zhang and Weijian Yi and Yulong Mao and Jinchen Liu and Yuzhuang Xu and Jinan Xu and Jian-Yun Nie and Yang Liu},year={2024},eprint={2405.10936},archivePrefix={arXiv},primaryClass={cs.CL},url={https://arxiv.org/abs/2405.10936}
}

@article{gallotta2024large,
  title={Large language models and games: A survey and roadmap},
  author={Gallotta, Roberto and Todd, Graham and Zammit, Marvin and Earle, Sam and Liapis, Antonios and Togelius, Julian and Yannakakis, Georgios N},
  journal={arXiv preprint arXiv:2402.18659},
  year={2024}
}

@article{latif2023sparks,
  title={Sparks of large audio models: A survey and outlook},
  author={Latif, Siddique and Shoukat, Moazzam and Shamshad, Fahad and Usama, Muhammad and Ren, Yi and Cuay{\'a}huitl, Heriberto and Wang, Wenwu and Zhang, Xulong and Togneri, Roberto and Cambria, Erik and others},
  journal={arXiv preprint arXiv:2308.12792},
  year={2023}
}

@article{tang2023video,
  title={Video understanding with large language models: A survey},
  author={Tang, Yunlong and Bi, Jing and Xu, Siting and Song, Luchuan and Liang, Susan and Wang, Teng and Zhang, Daoan and An, Jie and Lin, Jingyang and Zhu, Rongyi and others},
  journal={arXiv preprint arXiv:2312.17432},
  year={2023}
}

@article{zhang2023large,
  title={When Large Language Models Meet Citation: A Survey},
  author={Zhang, Yang and Wang, Yufei and Wang, Kai and Sheng, Quan Z and Yao, Lina and Mahmood, Adnan and Zhang, Wei Emma and Zhao, Rongying},
  journal={arXiv preprint arXiv:2309.09727},
  year={2023}
}

@article{hu2024rag,
  title={Rag and rau: A survey on retrieval-augmented language model in natural language processing},
  author={Hu, Yucheng and Lu, Yuxing},
  journal={arXiv preprint arXiv:2404.19543},
  year={2024}
}

@misc{zhao2023retrievingmultimodalinformationaugmented,
      title={Retrieving Multimodal Information for Augmented Generation: A Survey}, 
      author={Ruochen Zhao and Hailin Chen and Weishi Wang and Fangkai Jiao and Xuan Long Do and Chengwei Qin and Bosheng Ding and Xiaobao Guo and Minzhi Li and Xingxuan Li and Shafiq Joty},
      year={2023},
      eprint={2303.10868},
      archivePrefix={arXiv},
      primaryClass={cs.CL},
      url={https://arxiv.org/abs/2303.10868}, 
}

@article{xu2023large,
  title={Large language models for generative information extraction: A survey},
  author={Xu, Derong and Chen, Wei and Peng, Wenjun and Zhang, Chao and Xu, Tong and Zhao, Xiangyu and Wu, Xian and Zheng, Yefeng and Chen, Enhong},
  journal={arXiv preprint arXiv:2312.17617},
  year={2023}
}

@article{jin2023large,
  title={Large language models on graphs: A comprehensive survey},
  author={Jin, Bowen and Liu, Gang and Han, Chi and Jiang, Meng and Ji, Heng and Han, Jiawei},
  journal={arXiv preprint arXiv:2312.02783},
  year={2023}
}

@article{chen2024knowledge,
  title={Knowledge graphs meet multi-modal learning: A comprehensive survey},
  author={Chen, Zhuo and Zhang, Yichi and Fang, Yin and Geng, Yuxia and Guo, Lingbing and Chen, Xiang and Li, Qian and Zhang, Wen and Chen, Jiaoyan and Zhu, Yushan and others},
  journal={arXiv preprint arXiv:2402.05391},
  year={2024}
}

@article{li2023survey,
  title={A survey of graph meets large language model: Progress and future directions},
  author={Li, Yuhan and Li, Zhixun and Wang, Peisong and Li, Jia and Sun, Xiangguo and Cheng, Hong and Yu, Jeffrey Xu},
  journal={arXiv preprint arXiv:2311.12399},
  year={2023}
}

@misc{bai2024surveymultimodallargelanguage,
      title={A Survey of Multimodal Large Language Model from A Data-centric Perspective}, 
      author={Tianyi Bai and Hao Liang and Binwang Wan and Yanran Xu and Xi Li and Shiyu Li and Ling Yang and Bozhou Li and Yifan Wang and Bin Cui and Ping Huang and Jiulong Shan and Conghui He and Binhang Yuan and Wentao Zhang},
      year={2024},
      eprint={2405.16640},
      archivePrefix={arXiv},
      primaryClass={cs.AI},
      url={https://arxiv.org/abs/2405.16640}, 
}

@article{yang2024if,
  title={If llm is the wizard, then code is the wand: A survey on how code empowers large language models to serve as intelligent agents},
  author={Yang, Ke and Liu, Jiateng and Wu, John and Yang, Chaoqi and Fung, Yi R and Li, Sha and Huang, Zixuan and Cao, Xu and Wang, Xingyao and Wang, Yiquan and others},
  journal={arXiv preprint arXiv:2401.00812},
  year={2024}
}

@article{feng2023trends,
  title={Trends in integration of knowledge and large language models: A survey and taxonomy of methods, benchmarks, and applications},
  author={Feng, Zhangyin and Ma, Weitao and Yu, Weijiang and Huang, Lei and Wang, Haotian and Chen, Qianglong and Peng, Weihua and Feng, Xiaocheng and Qin, Bing and others},
  journal={arXiv preprint arXiv:2311.05876},
  year={2023}
}

@article{shi2024continual,
  title={Continual learning of large language models: A comprehensive survey},
  author={Shi, Haizhou and Xu, Zihao and Wang, Hengyi and Qin, Weiyi and Wang, Wenyuan and Wang, Yibin and Wang, Hao},
  journal={arXiv preprint arXiv:2404.16789},
  year={2024}
}

@misc{jin2024efficientmultimodallargelanguage,
      title={Efficient Multimodal Large Language Models: A Survey}, 
      author={Yizhang Jin and Jian Li and Yexin Liu and Tianjun Gu and Kai Wu and Zhengkai Jiang and Muyang He and Bo Zhao and Xin Tan and Zhenye Gan and Yabiao Wang and Chengjie Wang and Lizhuang Ma},
      year={2024},
      eprint={2405.10739},
      archivePrefix={arXiv},
      primaryClass={cs.CV},
      url={https://arxiv.org/abs/2405.10739}, 
}

@misc{liu2024fewshotadaptationmultimodalfoundation,
      title={Few-shot Adaptation of Multi-modal Foundation Models: A Survey}, 
      author={Fan Liu and Tianshu Zhang and Wenwen Dai and Wenwen Cai and Xiaocong Zhou and Delong Chen},
      year={2024},
      eprint={2401.01736},
      archivePrefix={arXiv},
      primaryClass={cs.CV},
      url={https://arxiv.org/abs/2401.01736}, 
}

@ARTICLE{wang23surveyonllmautoagents,
  title         = "A survey on large language model based autonomous agents",
  author        = "Wang, Lei and Ma, Chen and Feng, Xueyang and Zhang, Zeyu and
                   Yang, Hao and Zhang, Jingsen and Chen, Zhiyuan and Tang,
                   Jiakai and Chen, Xu and Lin, Yankai and Zhao, Wayne Xin and
                   Wei, Zhewei and Wen, Ji-Rong",
  journal       = "arXiv [cs.AI]",
  month         =  aug,
  year          =  2023,
  archivePrefix = "arXiv",
  primaryClass  = "cs.AI"
}

@ARTICLE{xie24lma,
  title         = "Large Multimodal Agents: A Survey",
  author        = "Xie, Junlin and Chen, Zhihong and Zhang, Ruifei and Wan,
                   Xiang and Li, Guanbin",
  journal       = "arXiv [cs.CV]",
  month         =  feb,
  year          =  2024,
  archivePrefix = "arXiv",
  primaryClass  = "cs.CV"
}

@article{niu2024large,
  title={Large Language Models and Cognitive Science: A Comprehensive Review of Similarities, Differences, and Challenges},
  author={Niu, Qian and Liu, Junyu and Bi, Ziqian and Feng, Pohsun and Peng, Benji and Chen, Keyu},
  journal={arXiv preprint arXiv:2409.02387},
  year={2024}
}

@article{devlin2018bert,
  title={Bert: Pre-training of deep bidirectional transformers for language understanding},
  author={Devlin, Jacob and Chang, Ming-Wei and Lee, Kenton and Toutanova, Kristina},
  journal={arXiv preprint arXiv:1810.04805},
  year={2018}
}

@article{radford2018improving,
  title={Improving language understanding by generative pre-training},
  author={Radford, Alec and Narasimhan, Karthik and Salimans, Tim and Sutskever, Ilya and others},
  year={2018},
  publisher={San Francisco, CA, USA}
}

@article{radford2019language,
  title={Language models are unsupervised multitask learners},
  author={Radford, Alec and Wu, Jeffrey and Child, Rewon and Luan, David and Amodei, Dario and Sutskever, Ilya and others},
  journal={OpenAI blog},
  volume={1},
  number={8},
  pages={9},
  year={2019}
}

@article{brown2020language,
  title={Language models are few-shot learners},
  author={Brown, Tom and Mann, Benjamin and Ryder, Nick and Subbiah, Melanie and Kaplan, Jared D and Dhariwal, Prafulla and Neelakantan, Arvind and Shyam, Pranav and Sastry, Girish and Askell, Amanda and others},
  journal={Advances in neural information processing systems},
  volume={33},
  pages={1877--1901},
  year={2020}
}

@article{frome2013devise,
  title={Devise: A deep visual-semantic embedding model},
  author={Frome, Andrea and Corrado, Greg S and Shlens, Jon and Bengio, Samy and Dean, Jeff and Ranzato, Marc'Aurelio and Mikolov, Tomas},
  journal={Advances in neural information processing systems},
  volume={26},
  year={2013}
}

@article{kiros2014unifying,
  title={Unifying visual-semantic embeddings with multimodal neural language models},
  author={Kiros, Ryan and Salakhutdinov, Ruslan and Zemel, Richard S},
  journal={arXiv preprint arXiv:1411.2539},
  year={2014}
}

@article{hardoon2004canonical,
  title={Canonical correlation analysis: An overview with application to learning methods},
  author={Hardoon, David R and Szedmak, Sandor and Shawe-Taylor, John},
  journal={Neural computation},
  volume={16},
  number={12},
  pages={2639--2664},
  year={2004},
  publisher={MIT Press}
}

@inproceedings{he2020momentum,
  title={Momentum contrast for unsupervised visual representation learning},
  author={He, Kaiming and Fan, Haoqi and Wu, Yuxin and Xie, Saining and Girshick, Ross},
  booktitle={Proceedings of the IEEE/CVF conference on computer vision and pattern recognition},
  pages={9729--9738},
  year={2020}
}

@inproceedings{chen2020simple,
  title={A simple framework for contrastive learning of visual representations},
  author={Chen, Ting and Kornblith, Simon and Norouzi, Mohammad and Hinton, Geoffrey},
  booktitle={International conference on machine learning},
  pages={1597--1607},
  year={2020},
  organization={PMLR}
}

@inproceedings{wu2018unsupervised,
  title={Unsupervised feature learning via non-parametric instance discrimination},
  author={Wu, Zhirong and Xiong, Yuanjun and Yu, Stella X and Lin, Dahua},
  booktitle={Proceedings of the IEEE conference on computer vision and pattern recognition},
  pages={3733--3742},
  year={2018}
}

@article{oord2018representation,
  title={Representation learning with contrastive predictive coding},
  author={Oord, Aaron van den and Li, Yazhe and Vinyals, Oriol},
  journal={arXiv preprint arXiv:1807.03748},
  year={2018}
}

@inproceedings{misra2020self,
  title={Self-supervised learning of pretext-invariant representations},
  author={Misra, Ishan and Maaten, Laurens van der},
  booktitle={Proceedings of the IEEE/CVF conference on computer vision and pattern recognition},
  pages={6707--6717},
  year={2020}
}

@article{caron2020unsupervised,
  title={Unsupervised learning of visual features by contrasting cluster assignments},
  author={Caron, Mathilde and Misra, Ishan and Mairal, Julien and Goyal, Priya and Bojanowski, Piotr and Joulin, Armand},
  journal={Advances in neural information processing systems},
  volume={33},
  pages={9912--9924},
  year={2020}
}

@article{grill2020bootstrap,
  title={Bootstrap your own latent-a new approach to self-supervised learning},
  author={Grill, Jean-Bastien and Strub, Florian and Altch{\'e}, Florent and Tallec, Corentin and Richemond, Pierre and Buchatskaya, Elena and Doersch, Carl and Avila Pires, Bernardo and Guo, Zhaohan and Gheshlaghi Azar, Mohammad and others},
  journal={Advances in neural information processing systems},
  volume={33},
  pages={21271--21284},
  year={2020}
}

@inproceedings{radford2021learning,
  title={Learning transferable visual models from natural language supervision},
  author={Radford, Alec and Kim, Jong Wook and Hallacy, Chris and Ramesh, Aditya and Goh, Gabriel and Agarwal, Sandhini and Sastry, Girish and Askell, Amanda and Mishkin, Pamela and Clark, Jack and others},
  booktitle={International conference on machine learning},
  pages={8748--8763},
  year={2021},
  organization={PMLR}
}

@inproceedings{dou2022empirical,
  title={An empirical study of training end-to-end vision-and-language transformers},
  author={Dou, Zi-Yi and Xu, Yichong and Gan, Zhe and Wang, Jianfeng and Wang, Shuohang and Wang, Lijuan and Zhu, Chenguang and Zhang, Pengchuan and Yuan, Lu and Peng, Nanyun and others},
  booktitle={Proceedings of the IEEE/CVF Conference on Computer Vision and Pattern Recognition},
  pages={18166--18176},
  year={2022}
}

@article{wang2021simvlm,
  title={Simvlm: Simple visual language model pretraining with weak supervision},
  author={Wang, Zirui and Yu, Jiahui and Yu, Adams Wei and Dai, Zihang and Tsvetkov, Yulia and Cao, Yuan},
  journal={arXiv preprint arXiv:2108.10904},
  year={2021}
}

@inproceedings{li2023blip,
  title={Blip-2: Bootstrapping language-image pre-training with frozen image encoders and large language models},
  author={Li, Junnan and Li, Dongxu and Savarese, Silvio and Hoi, Steven},
  booktitle={International conference on machine learning},
  pages={19730--19742},
  year={2023},
  organization={PMLR}
}

@inproceedings{rombach2022high,
  title={High-resolution image synthesis with latent diffusion models},
  author={Rombach, Robin and Blattmann, Andreas and Lorenz, Dominik and Esser, Patrick and Ommer, Bj{\"o}rn},
  booktitle={Proceedings of the IEEE/CVF conference on computer vision and pattern recognition},
  pages={10684--10695},
  year={2022}
}

@article{yang2023pixel,
  title={Pixel-aware stable diffusion for realistic image super-resolution and personalized stylization},
  author={Yang, Tao and Ren, Peiran and Xie, Xuansong and Zhang, Lei},
  journal={arXiv preprint arXiv:2308.14469},
  year={2023}
}

@article{podell2023sdxl,
  title={Sdxl: Improving latent diffusion models for high-resolution image synthesis},
  author={Podell, Dustin and English, Zion and Lacey, Kyle and Blattmann, Andreas and Dockhorn, Tim and M{\"u}ller, Jonas and Penna, Joe and Rombach, Robin},
  journal={arXiv preprint arXiv:2307.01952},
  year={2023}
}

@article{chen2020big,
  title={Big self-supervised models are strong semi-supervised learners},
  author={Chen, Ting and Kornblith, Simon and Swersky, Kevin and Norouzi, Mohammad and Hinton, Geoffrey E},
  journal={Advances in neural information processing systems},
  volume={33},
  pages={22243--22255},
  year={2020}
}

@article{chen2020improved,
  title={Improved baselines with momentum contrastive learning},
  author={Chen, Xinlei and Fan, Haoqi and Girshick, Ross and He, Kaiming},
  journal={arXiv preprint arXiv:2003.04297},
  year={2020}
}

@inproceedings{tian2020contrastive,
  title={Contrastive multiview coding},
  author={Tian, Yonglong and Krishnan, Dilip and Isola, Phillip},
  booktitle={Computer Vision--ECCV 2020: 16th European Conference, Glasgow, UK, August 23--28, 2020, Proceedings, Part XI 16},
  pages={776--794},
  year={2020},
  organization={Springer}
}

@book{jurafsky2000speech,
  title={Speech and Language Processing (3rd ed. draft)},
  author={Jurafsky, Dan and Martin, James H.},
  year={2024},
  url={https://web.stanford.edu/~jurafsky/slp3/ed3book.pdf},
  note={Draft chapters and slides available at: https://web.stanford.edu/~jurafsky/slp3/},
}

@article{grossberg2013recurrent,
  title={Recurrent neural networks},
  author={Grossberg, Stephen},
  journal={Scholarpedia},
  volume={8},
  number={2},
  pages={1888},
  year={2013},
}

@article{peters2018deep,
  author       = {Matthew E. Peters and
                  Mark Neumann and
                  Mohit Iyyer and
                  Matt Gardner and
                  Christopher Clark and
                  Kenton Lee and
                  Luke Zettlemoyer},
  title        = {Deep contextualized word representations},
  journal      = {CoRR},
  volume       = {abs/1802.05365},
  year         = {2018},
  url          = {http://arxiv.org/abs/1802.05365},
  eprinttype   = {arXiv},
  eprint       = {1802.05365},
  bibsource    = {dblp computer science bibliography, https://dblp.org},
}

@article{vaswani2017attention,
  title={Attention is all you need},
  author={Vaswani, Ashish and Shazeer, Noam and Parmar, Niki and Uszkoreit, Jakob and Jones, Llion and Gomez, Aidan N and Kaiser, {\L}ukasz and Polosukhin, Illia},
  journal={Advances in neural information processing systems},
  volume={30},
  year={2017},
}

@article{chung1967markov,
  title={Markov chains},
  author={Chung, Kai Lai},
  journal={Springer-Verlag, New York},
  year={1967},
  publisher={Springer},
}

@book{dynkin1965markov,
  title={Markov Processes},
  author={Dynkin, Evgeni{\u\i} Borisovich},
  year={1965},
  publisher={Springer},
}

@article{bahl1983maximum,
  title={A maximum likelihood approach to continuous speech recognition},
  author={Bahl, Lalit R and Jelinek, Frederick and Mercer, Robert L},
  journal={IEEE transactions on pattern analysis and machine intelligence},
  number={2},
  pages={179--190},
  year={1983},
  publisher={IEEE},
}

@article{hochreiter1997long,
  title={Long short-term memory},
  author={Hochreiter, Sepp and Schmidhuber, J{\"u}rgen},
  journal={Neural computation},
  volume={9},
  number={8},
  pages={1735--1780},
  year={1997},
  publisher={MIT press},
}

@article{lecun1998gradient,
  title={Gradient-based learning applied to document recognition},
  author={LeCun, Yann and Bottou, L{\'e}on and Bengio, Yoshua and Haffner, Patrick},
  journal={Proceedings of the IEEE},
  volume={86},
  number={11},
  pages={2278--2324},
  year={1998},
  publisher={Ieee},
}

@article{mikolov2013efficient,
  title={Efficient estimation of word representations in vector space},
  author={Mikolov, Tomas and Chen, Kai and Corrado, Greg and Dean, Jeffrey},
  journal={arXiv preprint arXiv:1301.3781},
  year={2013},
}

@inproceedings{pennington2014glove,
  title={Glove: Global vectors for word representation},
  author={Pennington, Jeffrey and Socher, Richard and Manning, Christopher D},
  booktitle={Proceedings of the 2014 conference on empirical methods in natural language processing (EMNLP)},
  pages={1532--1543},
  year={2014},
}

@article{sutskever2014sequence,
  title={Sequence to sequence learning with neural networks},
  author={Sutskever, Ilya and Vinyals, Oriol and Le, Quoc V},
  journal={Advances in neural information processing systems},
  volume={27},
  year={2014},
}

@article{bahdanau2014neural,
  title={Neural machine translation by jointly learning to align and translate},
  author={Bahdanau, Dzmitry and Cho, Kyunghyun and Bengio, Yoshua},
  journal={arXiv preprint arXiv:1409.0473},
  year={2014},
}

@article{sennrich2015neural,
  title={Neural machine translation of rare words with subword units},
  author={Sennrich, Rico and Haddow, Barry and Birch, Alexandra},
  journal={arXiv preprint arXiv:1508.07909},
  year={2015},
}

@article{krizhevsky2012imagenet,
  title={Imagenet classification with deep convolutional neural networks},
  author={Krizhevsky, Alex and Sutskever, Ilya and Hinton, Geoffrey E},
  journal={Advances in neural information processing systems},
  volume={25},
  year={2012},
}

@inproceedings{deng2009imagenet,
  title={Imagenet: A large-scale hierarchical image database},
  author={Deng, Jia and Dong, Wei and Socher, Richard and Li, Li-Jia and Li, Kai and Fei-Fei, Li},
  booktitle={2009 IEEE conference on computer vision and pattern recognition},
  pages={248--255},
  year={2009},
  organization={Ieee},
}

@article{DBLP:journals/corr/abs-2102-05918,
  author       = {Chao Jia and
                  Yinfei Yang and
                  Ye Xia and
                  Yi{-}Ting Chen and
                  Zarana Parekh and
                  Hieu Pham and
                  Quoc V. Le and
                  Yun{-}Hsuan Sung and
                  Zhen Li and
                  Tom Duerig},
  title        = {Scaling Up Visual and Vision-Language Representation Learning With
                  Noisy Text Supervision},
  journal      = {CoRR},
  volume       = {abs/2102.05918},
  year         = {2021},
  url          = {https://arxiv.org/abs/2102.05918},
  eprinttype    = {arXiv},
  eprint       = {2102.05918},
  bibsource    = {dblp computer science bibliography, https://dblp.org}
}

@article{DBLP:journals/corr/abs-2111-11432,
  author       = {Lu Yuan and
                  Dongdong Chen and
                  Yi{-}Ling Chen and
                  Noel Codella and
                  Xiyang Dai and
                  Jianfeng Gao and
                  Houdong Hu and
                  Xuedong Huang and
                  Boxin Li and
                  Chunyuan Li and
                  Ce Liu and
                  Mengchen Liu and
                  Zicheng Liu and
                  Yumao Lu and
                  Yu Shi and
                  Lijuan Wang and
                  Jianfeng Wang and
                  Bin Xiao and
                  Zhen Xiao and
                  Jianwei Yang and
                  Michael Zeng and
                  Luowei Zhou and
                  Pengchuan Zhang},
  title        = {Florence: {A} New Foundation Model for Computer Vision},
  journal      = {CoRR},
  volume       = {abs/2111.11432},
  year         = {2021},
  url          = {https://arxiv.org/abs/2111.11432},
  eprinttype    = {arXiv},
  eprint       = {2111.11432},
  bibsource    = {dblp computer science bibliography, https://dblp.org}
}

@article{DBLP:journals/corr/abs-1909-11740,
  author       = {Yen{-}Chun Chen and
                  Linjie Li and
                  Licheng Yu and
                  Ahmed El Kholy and
                  Faisal Ahmed and
                  Zhe Gan and
                  Yu Cheng and
                  Jingjing Liu},
  title        = {{UNITER:} Learning UNiversal Image-TExt Representations},
  journal      = {CoRR},
  volume       = {abs/1909.11740},
  year         = {2019},
  url          = {http://arxiv.org/abs/1909.11740},
  eprinttype    = {arXiv},
  eprint       = {1909.11740},
  bibsource    = {dblp computer science bibliography, https://dblp.org}
}

@misc{kim2021viltvisionandlanguagetransformerconvolution,
      title={ViLT: Vision-and-Language Transformer Without Convolution or Region Supervision}, 
      author={Wonjae Kim and Bokyung Son and Ildoo Kim},
      year={2021},
      eprint={2102.03334},
      archivePrefix={arXiv},
      primaryClass={stat.ML},
      url={https://arxiv.org/abs/2102.03334}, 
}

@misc{ge2024mllmbenchevaluatingmultimodalllms,
      title={MLLM-Bench: Evaluating Multimodal LLMs with Per-sample Criteria}, 
      author={Wentao Ge and Shunian Chen and Guiming Hardy Chen and Zhihong Chen and Junying Chen and Shuo Yan and Chenghao Zhu and Ziyue Lin and Wenya Xie and Xinyi Zhang and Yichen Chai and Xiaoyu Liu and Dingjie Song and Xidong Wang and Anningzhe Gao and Zhiyi Zhang and Jianquan Li and Xiang Wan and Benyou Wang},
      year={2024},
      eprint={2311.13951},
      archivePrefix={arXiv},
      primaryClass={cs.CL},
      url={https://arxiv.org/abs/2311.13951}, 
}

@misc{chen2024mllmasajudgeassessingmultimodalllmasajudge,
      title={MLLM-as-a-Judge: Assessing Multimodal LLM-as-a-Judge with Vision-Language Benchmark}, 
      author={Dongping Chen and Ruoxi Chen and Shilin Zhang and Yinuo Liu and Yaochen Wang and Huichi Zhou and Qihui Zhang and Yao Wan and Pan Zhou and Lichao Sun},
      year={2024},
      eprint={2402.04788},
      archivePrefix={arXiv},
      primaryClass={cs.CL},
      url={https://arxiv.org/abs/2402.04788}, 
}

@misc{pi2024mllmprotectorensuringmllmssafety,
      title={MLLM-Protector: Ensuring MLLM's Safety without Hurting Performance}, 
      author={Renjie Pi and Tianyang Han and Jianshu Zhang and Yueqi Xie and Rui Pan and Qing Lian and Hanze Dong and Jipeng Zhang and Tong Zhang},
      year={2024},
      eprint={2401.02906},
      archivePrefix={arXiv},
      primaryClass={cs.CR},
      url={https://arxiv.org/abs/2401.02906}, 
}

@misc{yao2024minicpmvgpt4vlevelmllm,
      title={MiniCPM-V: A GPT-4V Level MLLM on Your Phone}, 
      author={Yuan Yao and Tianyu Yu and Ao Zhang and Chongyi Wang and Junbo Cui and Hongji Zhu and Tianchi Cai and Haoyu Li and Weilin Zhao and Zhihui He and Qianyu Chen and Huarong Zhou and Zhensheng Zou and Haoye Zhang and Shengding Hu and Zhi Zheng and Jie Zhou and Jie Cai and Xu Han and Guoyang Zeng and Dahai Li and Zhiyuan Liu and Maosong Sun},
      year={2024},
      eprint={2408.01800},
      archivePrefix={arXiv},
      primaryClass={cs.CV},
      url={https://arxiv.org/abs/2408.01800}, 
}

@misc{ma2024eemllmdataefficientcomputeefficientmultimodal,
      title={EE-MLLM: A Data-Efficient and Compute-Efficient Multimodal Large Language Model}, 
      author={Feipeng Ma and Yizhou Zhou and Hebei Li and Zilong He and Siying Wu and Fengyun Rao and Yueyi Zhang and Xiaoyan Sun},
      year={2024},
      eprint={2408.11795},
      archivePrefix={arXiv},
      primaryClass={cs.CV},
      url={https://arxiv.org/abs/2408.11795}, 
}

@misc{chen2024mllmstrongrerankeradvancing,
      title={MLLM Is a Strong Reranker: Advancing Multimodal Retrieval-augmented Generation via Knowledge-enhanced Reranking and Noise-injected Training}, 
      author={Zhanpeng Chen and Chengjin Xu and Yiyan Qi and Jian Guo},
      year={2024},
      eprint={2407.21439},
      archivePrefix={arXiv},
      primaryClass={cs.AI},
      url={https://arxiv.org/abs/2407.21439}, 
}

@article{carolan2024review,
  title={A Review of Multi-Modal Large Language and Vision Models},
  author={Carolan, Kilian and Fennelly, Laura and Smeaton, Alan F},
  journal={arXiv preprint arXiv:2404.01322},
  year={2024}
}

\begin{figure}[htbp]
    \centering
    \includegraphics[width=\textwidth]{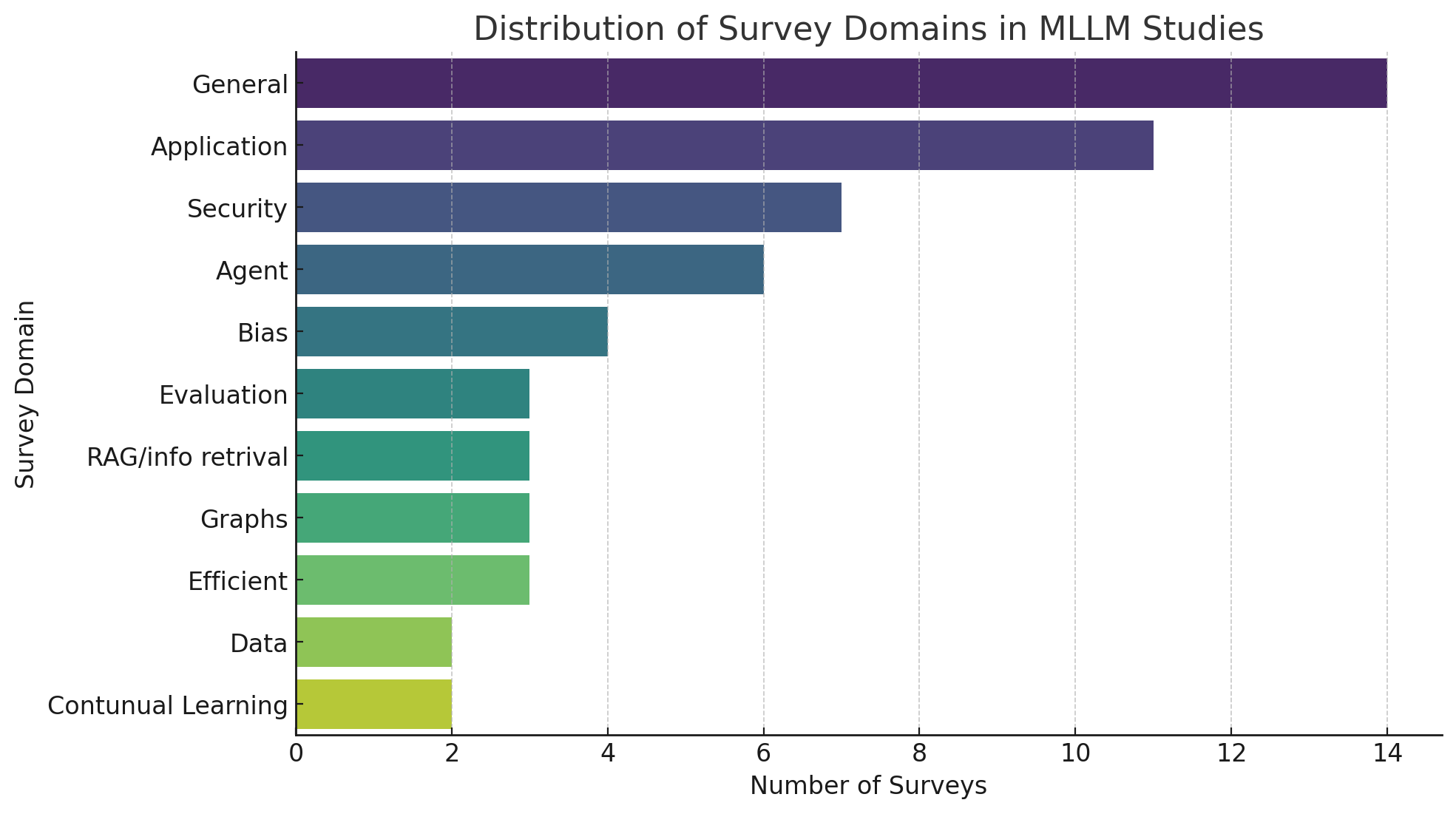}
    \caption{Distribution of Survey Domains in MLLM Studies}
\end{figure}

\begin{figure}[htbp]
    \centering
    \includegraphics[width=\textwidth]{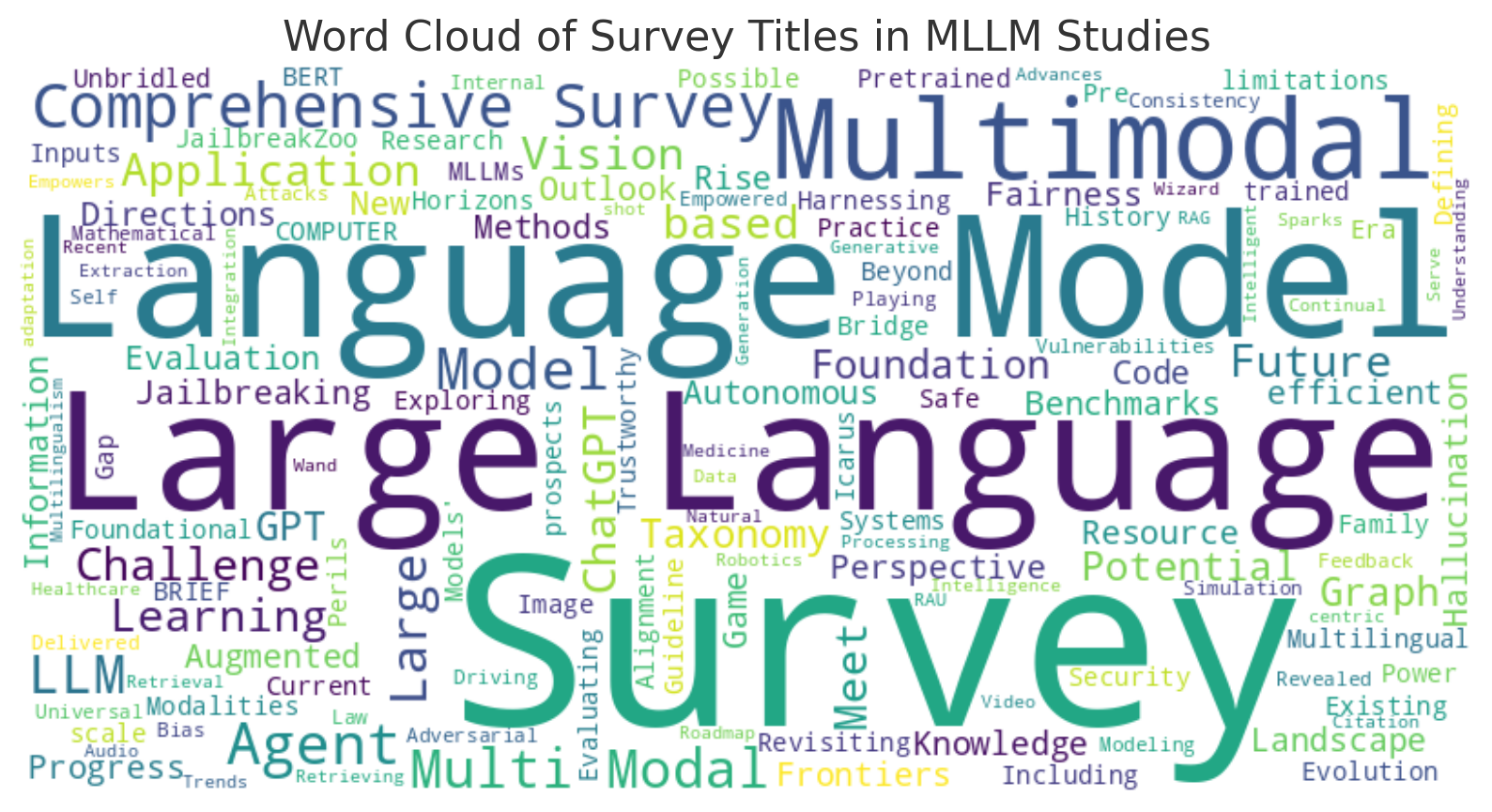}
    \caption{Word Cloud Of Survey Titles In MLLM Studies}
\end{figure}

\end{document}